\documentclass[11pt,journal,onecolumn]{IEEEtran}
\usepackage[left=1.25in, right=1.25in, top=1in, bottom=1in]{geometry}
\usepackage{setspace}
\usepackage[utf8]{inputenc}
\usepackage{amsmath,amsfonts,mathrsfs}
\usepackage[bookmarks=false]{hyperref}
\usepackage{longtable}
\usepackage{listings}
\usepackage{color}
\usepackage{enumitem}
\usepackage{soul}
\usepackage{diagbox}
\usepackage{graphicx}
\usepackage{textcomp}
\usepackage{multirow}
\usepackage{bm}
\usepackage{algorithm}
\usepackage[noend]{algorithmic}
\usepackage{subfig}
\usepackage{marvosym}
\usepackage{lineno}
 \usepackage{background}
\backgroundsetup{
	position=current page.north,
	angle=0,
	vshift=-13mm,
	opacity=1,
	scale=1,
	color=black!40,
	contents=Accepted at IEEE Aerospace and Electronic Systems Magazine 
}
\newcommand{\code}[1]{\lstinline!#1!}
\newcommand{\run}[1]{\medskip\noindent\lstinline!> #1!\medskip}
\newcommand{\runwin}[1]{\medskip\noindent\lstinline!WIN> #1!\medskip}

\newcommand{\supp}[1]{\textsuperscript{#1}}

\newcommand{\deltaT}{\Delta t}
\newcommand{\instancesLeaf}{m_{\text{LEAF}}}
\newcommand{\dataset}{\mathscr{D}}
\newcommand{\datatrain}{\dataset_\text{TRAIN}}
\newcommand{\datatest}{\dataset_\text{TEST}}

\newcommand{\importance}{\mathit{importance}}
\newcommand{\forest}{\mathcal{RF}}
\newcommand{\tree}{\mathcal{T}}
\newcommand{\node}{\mathscr{N}}
\newcommand{\leaf}{\mathscr{L}}
\newcommand{\features}{\mathcal{F}}
\newcommand{\rmse}{\mathit{RMSE}}
\newcommand{\armse}{\overline{\rmse}}

\newcommand{\xgb}{{\it XGB}}
\newcommand{\rfmtr}{{\it G-RF}}
\newcommand{\rfstr}{{\it L-RF}}

\makeatletter
\newcommand{\ALOOP}[1]{\ALC@it {\bf{where}} #1%
    \begin{ALC@loop}}
\newcommand{\ENDALOOP}{\end{ALC@loop}}%

\makeatother

\definecolor{energy}{RGB}{102,255,255}
\definecolor{dmop}{RGB}{255,102,102}
\definecolor{ftl}{RGB}{204,0,255}

\definecolor{darkpastelgreen}{rgb}{0.01, 0.75, 0.24}
\definecolor{darkorange}{rgb}{1.0, 0.55, 0.0}

\setlist[enumerate,1]{%
	label=\arabic*.,
}
\newlist{inlinelist}{enumerate*}{1}
\setlist*[inlinelist,1]{%
	label=(\roman*),
}

\begin{document}
\doublespacing
%\title{Predicting thermal power consumption of the Mars Express satellite with machine learning}

\title{Machine learning for predicting thermal power consumption of the \\Mars Express Spacecraft}

\author{\IEEEauthorblockN{Matej Petkovi\'{c}\supp{1,2,\Letter}\footnote{\Letter{} Corresponding author\\
        Jo\v{z}ef Stefan Institute, Jamova Cesta 39, 1000 Ljubljana, Slovenia\\
        phone: +386 1 477 3635}, Redouane Boumghar\supp{3}, Martin Breskvar\supp{1,2}, Sa\v{s}o D\v{z}eroski\supp{1,2}, \\Dragi Kocev\supp{1,2}, Jurica Levati\'{c}\supp{4}, Luke Lucas\supp{5}, Alja\v{z} Osojnik\supp{1}, Bernard \v{Z}enko\supp{1},\\ Nikola Simidjievski\supp{1}\\}
		\IEEEauthorblockA{%
		\small
        e-mail: name.surname@ijs.si\supp{1}\\
        Jo\v{z}ef Stefan Institute, Slovenia\supp{1} \\
        Jo\v{z}ef Stefan International Postgraduate School, Slovenia\supp{2}\\
        e-mail: redouane.boumghar@free.fr\supp{3}\\
        Data Analytics Team for Operations, ESOC, European Space Agency, Germany\supp{3}\\
        e-mail: jurica.levatic@irbbarcelona.org\supp{4}\\
        Institute for Research in Biomedicine, Spain\supp{4}\\
        e-mail: luke.lucas@lsespace.com\supp{5}\\
        Mars Express, Mission Planning \& Spacecraft Operations, Germany\supp{5}\\
        }
}


\maketitle

% \fi
\IEEEpeerreviewmaketitle

%\color{red}
%\color{red}
\begin{abstract}
The thermal subsystem of the Mars Express (MEX) spacecraft keeps the on-board equipment within its pre-defined operating temperatures range. To plan and optimize the scientific operations of MEX, its operators need to estimate in advance, as accurately as possible, the power consumption of the thermal subsystem. The remaining power can then be allocated for scientific purposes. We present a machine learning pipeline for efficiently constructing accurate predictive models for predicting the power of the thermal subsystem on board MEX. In particular, we employ state-of-the-art feature engineering approaches for transforming raw telemetry data, in turn used for constructing accurate models with different state-of-the-art machine learning methods. We show that the proposed pipeline considerably improve our previous (competition-winning) work in terms of time efficiency and predictive performance. Moreover, while achieving superior predictive performance, the constructed models also provide important insight into the spacecraft's behavior, allowing for further analyses and optimal planning of MEX's operation. 
\end{abstract}

\begin{IEEEkeywords}
machine learning, Mars Express spacecraft, ensemble learning, predictive modeling, random forest, gradient boosting, feature engineering
\end{IEEEkeywords}

% \fi
\IEEEpeerreviewmaketitle

%\linenumbers
\section{Introduction}
\IEEEPARstart{M}{ars Express (MEX)}, a spacecraft operated by the European Space Agency (ESA), is Europe’s first spacecraft that orbits Mars. During its science operations, since the beginning of 2004, it has provided evidence of the presence of water above and below the surface of the planet \cite{Oroseieaar7268}, an ample amount of three-dimensional renders of the surface as well as the most complete map of the chemical composition of Mars’s atmosphere \cite{chicarro2004mars}.

MEX is powered by electricity generated by its solar arrays and stored in batteries to be used during the eclipse periods. The scientific payload of the MEX consists of seven instruments that provide global coverage of the planet’s surface, subsurface and atmosphere. The instruments and on-board equipment have to be kept within their operating temperature ranges, spanning from room temperature for some instruments, to temperatures as low as –180\textdegree{}C for others. In order to maintain these predefined operating temperatures, the spacecraft is equipped with an autonomous thermal system composed of $33$ heater lines as well as coolers. The thermal system, together with the platform units, consumes a significant amount of the total generated electric power, leaving a fraction to be used for science operations.

\begin{figure*}[!b]
\centering
 \includegraphics[width=\textwidth]{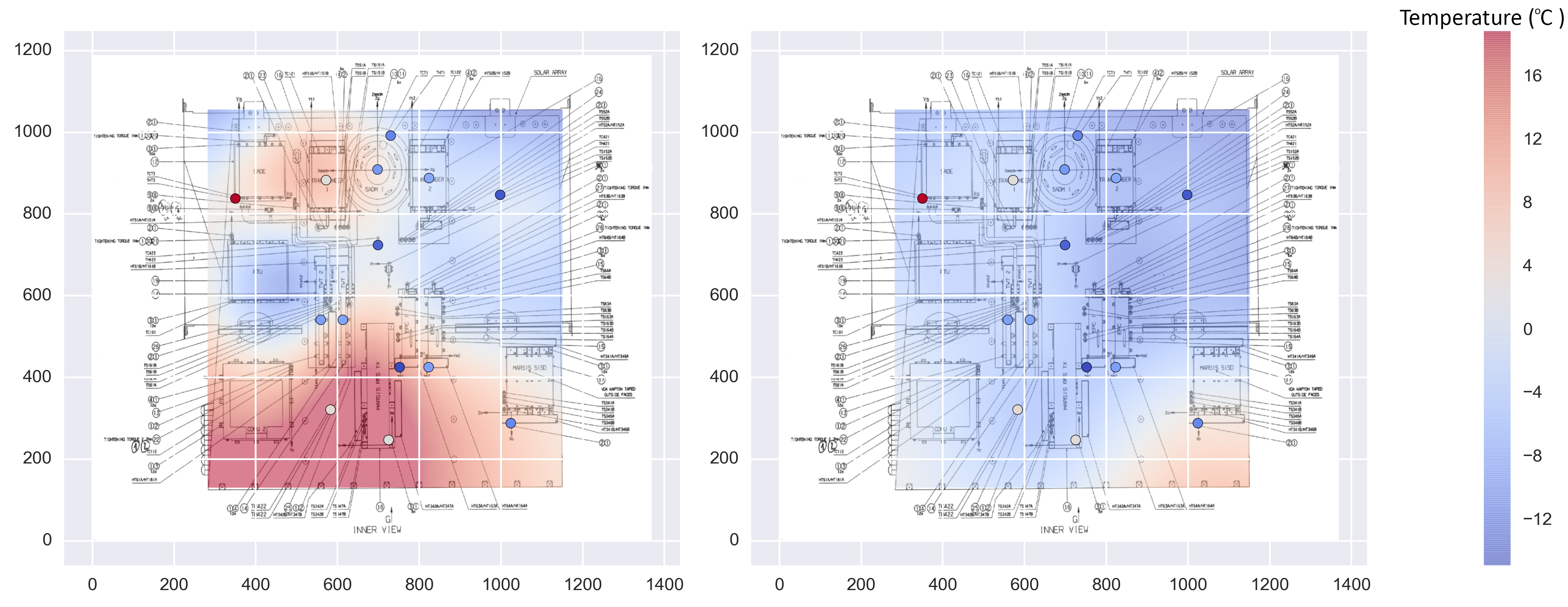}
\caption{ Interpolated thermal effects on the MEX's $-Y$ face, with radio transmitter turned on (left) and off (right)}
\label{fig:heatM}
\end{figure*}

Predicting the power consumption of the thermal system is a non-trivial task. However, due to the aging of the spacecraft and the decaying capacity of its batteries, it is a very crucial one for optimal planning and execution of science operations on MEX. The power consumption is a dynamic process that changes through time, depending on various external and internal factors, such as long-term exposure of the spacecraft to the Sun or heat generated by the on-board instruments. For instance, Figure~\ref{fig:heatM} shows the effect of the radio transmitter during a communication pass, with interpolation between different temperature sensors of the $-Y$ face of the spacecraft. Temperatures fluctuate by up to 28\textdegree{}C due to these two different on/off conditions. Current attempts at modeling and predicting the power consumption, involve manually constructed models that are based on simplified first-principle models, expert knowledge and experience. Given MEX's current condition, this prompts for a more accurate predictive model of the thermal power consumption, which would yield prolonged operating life.

This motivated the organization of the first ESA’s data mining competition – the Mars Express Power Challenge \cite{lucas2017:mars}. The focus of the challenge was the development of specialized approaches for constructing models that are able to accurately estimate and predict the MEX’s thermal power consumption (TPC) given only measured telemetry data. For this task of predictive modeling, machine learning approaches offer a different, yet more accurate solution to modeling the complex relationship between the telemetry and the power consumption than a human expert.

Machine learning is an area in the realm of artificial intelligence, which studies algorithms with the ability to learn, i.e., algorithms that improve their performance through knowledge gathered from experience (data). Their ability to capture and describe patterns in complex data makes them a valuable asset for studying a variety of phenomena in different domains from life sciences, earth sciences, social and behavioral sciences. In the context of the MEX challenge, machine learning algorithms for predictive modeling aim at constructing a model that can capture complex relationships in the data. In turn, such models accurately estimate future values of power consumption for each of the $33$ heater lines and coolers (target variables/features) given measured telemetry data (descriptive variables/features).

In our previous work \cite{smc:2017}, we presented the machine learning pipeline solution that won the Mars Express Power Challenge. The winning solution first transforms the raw telemetry data into carefully constructed features with $1$ minute time resolution between values, rendering a massive data set. Next, it uses the method of Random Forest of Predictive Clustering Trees (RF-PCTs) \cite{Kocev13:jrnl} to construct $33$ predictive models for each of the $33$ target variables. Finally, it outputs a predicted value of each target variable for every hour of one Martian year in the future ($1.88$ Earth years). The proposed solution performed better than the $\sim$40 other competing solutions while being more accurate than the models currently in use at ESA by an order of magnitude.

However, the premium predictive accuracy of the winning solution came at a cost of substantial computational overhead. In this paper, we extend the work presented in \cite{smc:2017} to address this issue. In particular, we propose an update to the winning solution which aims at efficiently constructing predictive models of MEX's TPC, while still being able to maintain good predictive performance. More specifically, we consider updates of the pipeline along two dimensions: (1) constructing data with different data granularity in the learning process and (2) using different machine learning methods which can efficiently learn accurate predictive models. The former considers engineering features from the raw telemetry data at different time resolutions coarser than $1$ minute thus reducing the size of the data set used in the learning phase. The latter considers both local and global methods for multi-target regression. Local methods construct a model for each target variable separately. Here, besides the winning method RFs of PCTs \cite{Kocev13:jrnl}, we also consider XGBoost \cite{Chen2016}, a recent efficient implementation of Stochastic Gradient Boosting \cite{Friedman2002}. In contrast, global methods produce a predictive model able to predict several target variables simultaneously \cite{borchani}. To this end, we consider global RFs of PCTs for multi-target regression, an extension of the local version, which can construct single model for all $33$ target variables, therefore substantially reducing the computational time needed for obtaining a solution \cite{Kocev13:jrnl,Spyromitros-Xioufis2016}. 

In sum, the main task that we address is: Given three Martian years of telemetry data (August 22, 2008 to April 14, 2014), use machine learning to efficiently construct predictive model that accurately predicts the values of the electric current through the $33$ thermal power consumers for the subsequent Martian year (April 14, 2014 to March 1, 2016).

The remainder of the paper is organized as follows. In Section~2, we provide an overview of the work related to machine learning applications for space-exploration research. Section~3 presents and discusses the tasks of data preparation and pre-processing. Section~4 presents the machine learning methods used in this study. In Section~5, we present the experimental setup for evaluating the proposed extensions of the machine learning pipeline. Section~6 presents and discusses the results of the empirical evaluation. Finally, Section~7 concludes the paper and suggests directions for further work.

\section{Related Work}

Machine learning offers an ample amount of methods that tackle predictive tasks in real life domains \cite{hastie2013elements,Witten2005}. These methods have been applied for predicting discrete output values (classification), continuous output values (regression), even structured outputs as in gene networks, image classification, text categorization etc \cite{Bakir2007}.

The challenges typically addressed in space-exploration research are associated with high-cost of failure \cite{McGovern2011}. For instance, the remote spacecraft are typically equipped with processors and memory lagging decades behind the state of the art. Next, the development and launch of a space mission is expensive and there is little or no opportunity for repair. In this context, the utility of machine learning approaches has been proven to be valuable asset. In particular, many applications of machine learning address the task of anomaly detection in spacecraft, where a typical task considers monitoring the status of the on-board equipment. Such analyses of telemetry data are performed using neural networks \cite{Zhenping:smc}, relevance vector machine \cite{Yairi} or by applying seasonal decomposition methods (linear regression together with the nearest neighbours)\cite{Munoz:smc}. 

Machine learning can be used not only for estimating the current state of a spacecraft, but also predicting its future ones and therefore allowing for autonomous decisions. In our previous work \cite{smc:2017}, we propose a solution for predicting spacecraft's power consumption, which can be used for optimizing its operation. This works closely relates to the one of \cite{canio:spaceops}, where the authors propose Random Forests for predicting temperature of the instruments to optimize battery usage during eclipses. 

Another important challenge relates to safe ground movement of autonomous (space) rovers or robots. The authors of \cite{Hernandez:smc} address this issue by using support vector machines to recognize and avoid dangerous objects. In similar context, \cite{Giusti:forest-paths} addresses the task of image analysis for automated navigation systems. Here, with deep neural Networks as the underlying machine learning method, autonomous drones are utilized for tracking forests paths.

Finally, machine learning can be also utilized to learn or simplify a physical model of a spacecraft or a model of its environment. In \cite{finn:spaceops}, the authors employ Random Forests to simplify the exact physical model for complex and dynamic radiation environment in the Van Allen belts. In a similar context, \cite{McGovern2011} outlines several studies which address challenges of spacecraft operating in high-radiation environments (beyond Earth's magnetosphere and ionosphere) and the reliability of machine learning algorithms applied in these scenarios. In particular, these studies propose variants of traditional ML approaches (k-means and SVMs) robust to potential data corruption on the disks due to the various levels of radiation.

\section{The Data}\label{sec:feats}

In the typical machine learning setting, the input in a learning algorithm is (training) data which embodies the experience. The data consists of training examples (also referred to as instances or measurements) and their properties (also referred to as features or attributes). The features, numerical (i.e., continuous) or nominal (i.e., discrete), can either describe the data or specify the desired output of the algorithm. In the context of predicting MEX's TPC, an example is a time period, while features are derived from context and observations data. 

In this paper, we use data provided by ESA \cite{lucas2017:mars}, that consists of raw telemetry data (context data) and measurements of the electric current of $33$ thermal power lines (observation data), for three Martian years of MEX operations. We refer to these data as the training data $\datatrain$. For the fourth Martian year of the operation, the context part of the data was used for generating the predicted values, that in turn where evaluated using the real measured observation data. We refer to these data as the test data $\datatest$. 

The \emph{observation data} consists of the electric current measurements of the 33 power consumers, recorded once or twice per minute. The \emph{context data} consists of five components:
	
\begin{itemize}
\item {\bf SAA} (Solar Aspect Angles) data contain the angles between the Sun--MEX line and the axes of the MEX's coordinate system.
	
\item {\bf DMOP} (Detailed Mission Operations Plans) data contain the information about the execution of different subsystems' commands at a specific time.
	
\item {\bf FTL} (Flight dynamics TimeLine events) data contain the pointing and action commands that impact the position of MEX, such as pointing the spacecraft towards Earth or Mars.
	
\item {\bf EVTF} (Miscellaneous Events) data contain time intervals during which MEX was in Mars's shadow or records of the time points when the MEX is in apsis of its elliptical orbit.
	
\item {\bf LTDATA} (Long Term Data) contain the Sun--Mars distances and the solar constant.
	
\end{itemize}

All raw data entries are time-stamped (expressed in milliseconds) indicating when the entry was logged. The time span between the two consecutive entries varies from less than a minute (SAA) to several hours (LTDATA). For a detailed description of the task and the data, we refer the reader to \cite{lucas2017:mars,boumghar:spaceops}.

The raw data is not directly applicable to a machine learning algorithm due to two main reasons: (i) incompatible time resolutions of the different components of the raw data, and (ii) unstructured format of some of the entries, such as text, that are not readily usable for machine learning algorithms. Therefore, to construct an appropriate data set for a learning algorithm, we pre-process the raw data in two phases: (i) conveying data time resolution (time interval between two consecutive examples) and (ii) engineering new (more informative) features from different parts of the context data that may yield to better predictive performance.

The first phase relates to choosing an appropriate time resolution $\deltaT$ of the data set, and divide the time span $[t_\text{FIRST}, t_\text{LAST})$ into subintervals $[t_i, t_{i + 1})$, where $t_{i + 1} = t_i + \deltaT$. Here, $t_\text{FIRST}$ is the first time-stamp in the $\datatrain$, and $t_\text{LAST}$ is the last time-stamp in the $\datatest$.

The second phase considers constructing more informative features. The value of a given feature for a particular time interval is obtained by aggregating measurements from the time interval at hand that correspond to one or more components from the raw telemetry data.

Due to issues with the spacecraft communication in some periods, some measurements are missing from the both the context and observations data. In principle, the machine learning methods employed in this study can handle data with missing values. However, longer periods with contiguous missing values can substantially damage the accuracy of the learned predictive models as well as add an additional computational overhead. For this reason, we remove examples with missing observation data for time periods longer than $10$ minutes. On the other hand, in the context data, we interpolate the examples with missing values for time periods shorter than $10$ minutes, or leave them intact otherwise.  

In the following subsections, we describe the groups of features constructed in the pre-processing step of the pipeline.

\subsection{Energy influx features}

There are seven features in this group: one for solar panels and one for each of the six sides of the cuboid of MEX. The features describe the amount of solar energy that is collected through a given surface in a given time interval $[t_i, t_{i + 1})$. The solar energy collected by a side of
cuboid directly influence the amount of the energy used by the thermal lines that maintain the temperature in that part of the spacecraft. The solar energy collected by the solar panels influence the amount of available energy that can be stored in spacecraft’s batteries.

\begin{figure}[!ht]
\centering
\includegraphics[scale=0.20]{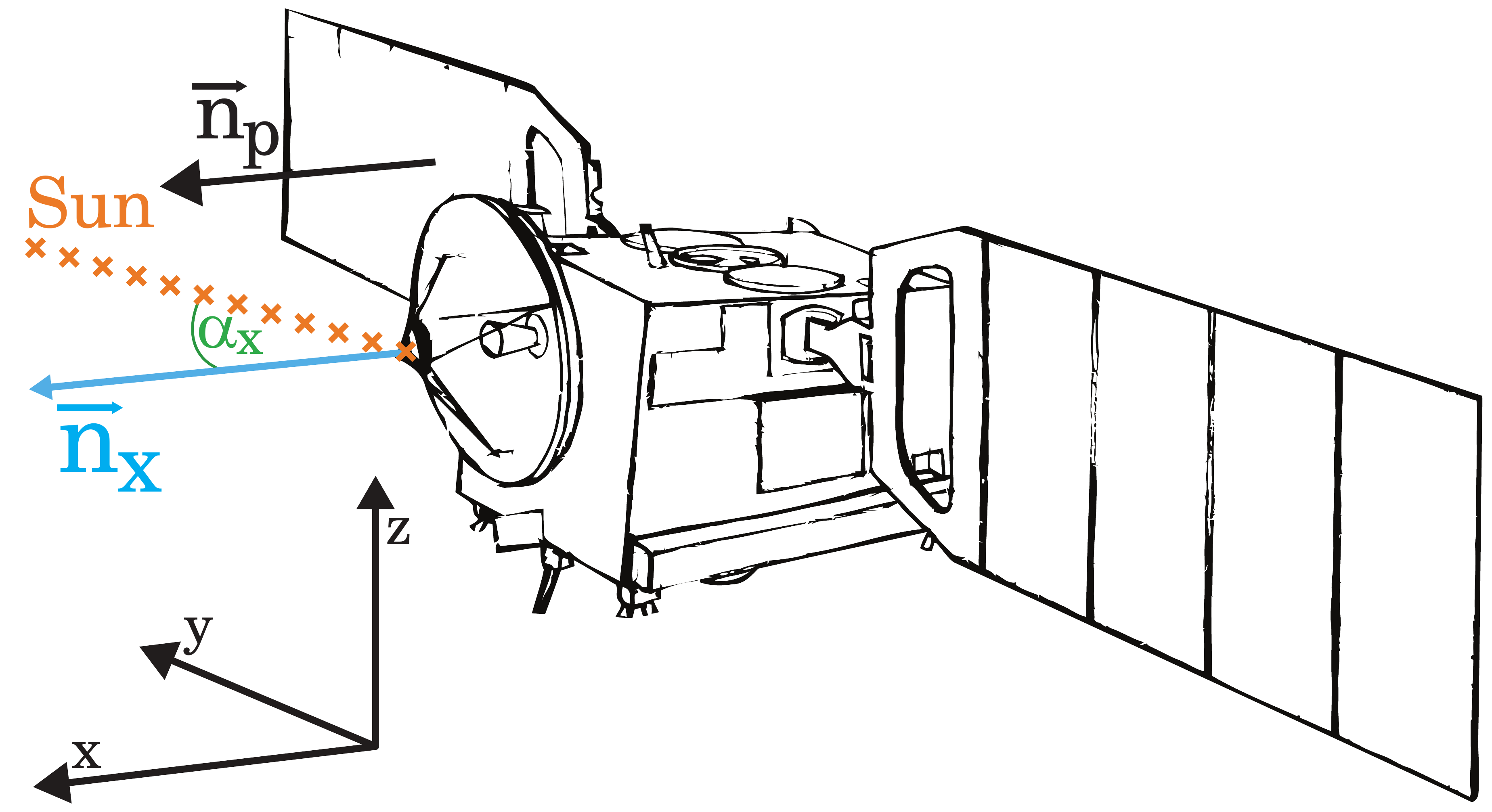}
\caption{Illustration of the MEX spacecraft and its coordinate axes $x$, $y$ and $z$, that correspond to \emph{front}, \emph{left} and \emph{up} sides of MEX, respectively. $\alpha_x$ denotes the solar aspect angle of the front side, i.e., the angle between the normal $\vec{n}_x$ and the Sun-MEX line. $\vec{n}_p$ denotes the normal of the panels.}
\label{fig:energy-influx-features}
\end{figure}

The amount of energy collected by a given surface is proportional to the product of the effective area $A_{\text{eff}}$ of the surface and the solar constant $c$. If the area $A$ is given, we compute $A_\text{eff}$ as $A_{\text{eff}} = A\max\{\cos\alpha, 0\}$\text{,} where $\alpha$ is the angle between the Sun--MEX line and the outer normal $\vec{n}$ to the surface (see Figure\ \ref{fig:energy-influx-features}). Without any loss of generality, we assume $A = 1$ for all surfaces, as the machine learning methods that we use, are invariant to monotonic transformations of features. The values of $\alpha$ for each of the seven surfaces were computed directly from the SAA data, while $c$ was given in LTDATA. In addition to the the effective area and the solar constant, (pen)umbras have a considerable impact on the energy influx. We define the amount of the energy $E_S^i$ that pass through the surface $S$ at time interval $[t_i, t_{i + 1})$ as
\[E_S^i = \int_{t_i}^{t_{i + 1}} A_\text{eff}(t)c(t)U(t)\mathrm{d}t\text{,}\]
where $U$ is the \emph{umbra coefficient}, an approximation of the proportion of Sun visible from the spacecraft. $U$ takes the value $U(t) = 0$ if the spacecraft is in an umbra, $U(t) = 0.5$ if the spacecraft is in a penumbra, and $U(t) = 1$ otherwise. Instead of calculating exact integrals for $E_S^i$, we approximate the values using the trapezoid-rule.

\subsection{Historical energy influx features}

The thermal state of the spacecraft depends not only on the current energy influx, but also on the energy influx in the past. To capture this, we construct historical energy influx features for each of the seven surfaces. A given historical feature for surface $S$ at time $t_i$ is computed as a sum of energy influx during given historical time-frame: 
\begin{equation}\label{eqn:historic-feats}
H_S^i = \sum_{j=1}^H E_S^{i-(j-1)}\text{,}
\end{equation}
where $H$ is the number of time-intervals included in the historical feature.
To account for different impacts of the historical energy influx we construct historical features with different time-frames,
for different values of $H$, given in Sec.~\ref{sec:paramethers}.

\subsection{DMOP features}

The DMOP data contain log of commands issued to different MEX sub-systems. The names of commands have been obfuscated, however the available documentation reveals two variants of events: (1) events that contain information about the subsystem and command that has been executed (e.g., ASXX383C) and (2)  events that represent flight dynamics events (e.g., MAPO.000005). 

The first four characters of the first variant represent the subsystem while the rest represent the command and its parameters. In the second variant the first four characters represent the name of the event, followed by a number that indicates the number of occurrences. Given that these events have different impact on the temperatures of various subsystems of the spacecraft, it is safe to assume that they impact the thermal subsystems differently.

More specifically, we assume that there is a significant delay between triggering of a sub-systems’ command and its actual effect on the thermal state of the spacecraft. Therefore, from the raw DMOP data, we construct features that encode this information of delayed effect in terms of ”time since last activation” of a specific sub-system command.

The values of the DMOP features are calculated as follows: 
\[f_{k}^{i}=
\begin{cases}
0 & \text{if $k$ is activated at $t_i$} \\
\min(f_{k}^{i-1} + \deltaT, \theta) & \text{otherwise}
\end{cases}\text{,}
\]
where $f_{k}^{i}$ denotes the value of feature corresponding to event $k$ at time $t_i$. Note that, here we also assume that all of the subsystems were deactivated at the first time point (i.e., $f_{k}^{0} = \theta$). The  $\theta$ regulates the effect of a given event diminishing with time, rendering its influence unimportant at some point. We selected this threshold to be  1 day ($\theta = 1440$, the number of minutes in a day). Table \ref{tab:dmop-features} presents these calculations of features.

\begin{table}[htb]
\centering
\caption{Illustration of the DMOP features that encode the time since last activation of a given subsystem command.}\label{tbl:dmop}
\label{tab:dmop-features}
\tabcolsep=0.11cm
\begin{tabular}{cc||cccc}
\multicolumn{2}{c}{Raw data} & \multicolumn{4}{c}{DMOP features} \\ \hline
t & Command & APSF28A1 & ASXX383C & ATTTF030A & ASXX303A \\
\hline
$t_1$ & \textit{none} & 1440 & 1440 & 1440 & 1440\\
$t_2$ & \textbf{ASXX383C} &  1440 & \textbf{0} & 1440 & 1440 \\
$t_3$ & \textbf{ATTTF030A} &   1440 & $\deltaT$ & \textbf{0} & 1440 \\
\vdots & \vdots & \vdots  & \vdots & \vdots & \vdots \\
$t_{14}$ & \textbf{ASXX303A} & 1440 & $12\deltaT$ & $11\deltaT$ & \textbf{0} \\
\end{tabular}
\end{table}

We construct such features for each flying dynamic event ($17$ features), each subsystem-command pair ($345$ features) and each sub-system in case different commands are issued to it ($15$ features). We also construct binary indicators for each subsystem and flying dynamic event ($34$ features in total), where a feature $f_{k}^{i}$ has value of 1 if the subsystem was triggered within the time-step $t_i$, and 0 otherwise.

\subsection{FTL features}

The FTL data contain logs of pointing events and their time ranges, 
where simultaneous events are also possible. For each pointing event in time interval $[t_i, t_{i + 1})$, a feature has value that equals the proportion of the time in $[t_i, t_{i + 1})$, during which the event is in progress. Since the duration of events is typically longer than
$\deltaT$, the values of the features are mostly 1 (event is in progress), or 0 (event not in progress). This approach renders 23 FTL features in total.

\subsection{The final data sets}

Table \ref{tab:datasets} presents the important details regarding the final constructed data sets used further in the experiments\footnote{The data are accessible at \url{http://spacelab.ijs.si/}}.

\begin{table}[!htb]
\centering
\caption{The properties of the final data sets given different granularity $\deltaT$.}
\label{tab:datasets}
\begin{tabular}{c | r r r }
 $\deltaT$ [min] & number of examples & number of features &  size [MB] \\
 \hline
 1 & 3922895 & 459 & 10090 \\
5  &  784773 & 459 & 2048 \\
10 &  392474 & 459 & 1045 \\
15 &  261697 & 459 & 702 \\
30 &  130900 & 459 & 353 \\
60 &   65493 & 445 & 168
\end{tabular}
\end{table}

\section{Machine learning methods}

Considering predictive modeling, many machine learning methods struggle with the problem of overfitting. Overfitting occurs when a method learns a model with a very good performance on the provided training data, but has limited generalization power and performs poorly on data unseen during learning. In machine learning, there is a long tradition of developing methods that address this problem of, by learning multiple (diverse) models and combining their outputs instead of just learning a single model. These methods are referred to as ensemble methods or ensembles. An ensemble is a set of (base) predictive models constructed with a given algorithm, that is expected to lead to predictive performance gain over an individual model by combining the predictions of its constituents. In this study, we employ two types of ensembles: (i) Random Forest of Predictive Clustering Trees (both local and global version of the algorithm for multi-target regression) \cite{Kocev13:jrnl,Breiman01:jrnl} and (ii) Stochastic Gradient Boosted Trees (XGBoost) \cite{Chen2016,Friedman2002}.

\subsection{Random Forest of Predictive Clustering Trees (PCTs)}

Random Forest (RF) \cite{Breiman01:jrnl} is an ensemble method that learns a set of tree-based predictive models and combines their prediction. The base models are learned from random bootstrap samples of the training set, where for each tree at each tree node a random feature subset (with a user defined size) is considered for selecting the best split. Such approach allows for constructing a set of diverse predictive models that can differ both in size and performance. 

In this study, the Random Forest ensembles consist of Predictive Clustering Trees (PCTs) \cite{blockeel-top-down-proc-1998}. PCTs have a tree structure that includes internal nodes and leaves. The internal nodes contain tests on the descriptive variables (i.e., the different features extracted with pre-processing), while leaves give predictions for the target variable (i.e., power consumption of a thermal line). PCT refers to a hierarchy of clusters with each node corresponding to a cluster. In particular, the top-node of a PCT corresponds to one cluster (group) containing all data points. This cluster is then recursively partitioned into smaller clusters while moving down the tree. The leaves represent the clusters at the lowest level of the hierarchy and each leaf is labeled with its cluster’s centroid/prototype (the average of the target variable is the prediction made by the leaf).

Random Forest of PCTs is a generalization of the traditional Random Forest ensemble of regression trees \cite{Breiman84:book}, in terms of addressing structured output prediction tasks \cite{Kocev13:jrnl,Blockeel98:phd}. While the traditional RF is able to predict values of a single numeric target variable at a time (i.e., is a local method for Multi-Target Regression), the RF of PCTs ensemble allows also for predicting several target variables simultaneously (i.e., is a global method for Multi-Target Regression). The algorithm for learning a RF of PCTs is presented in Alg.~\ref{alg:pct}. Namely, it takes three inputs: (1) training data $\datatrain$, as well as two hyper-parameters denoting (2) the number of trees $M$ in the ensemble and (3) the size of the feature subset considered at each node split $f$. Each PCT in the ensemble is learned with  greedy recursive top-down induction algorithm on a random bootstrap sample of the input data set (line \ref{algline:bSamle} of the Alg.~\ref{alg:pct})\footnote{The RF of PCT framework is implemented in the \textsc{CLUS} system available at \url{http://source.ijs.si/ktclus/clus-public}}. 

The PCT-induction algorithm (lines \ref{algline:pctS}-\ref{algline:pctEND} of the Alg.~\ref{alg:pct}) starts in the root node of the tree by selecting a test from set of candidate tests that are generated by a random feature subset $\features$ of size $f$. The best test is greedily chosen by a heuristic function that typically measures the impurity of the target values of the examples in the subsets $E_i$ of the set $E$ that this test results in. The goal of the heuristic function is to guide the algorithm towards small trees with good predictive performance. In the global MTR setting, this heuristic function is the average variance of the targets. In our case, with numeric features, every test is of form $x_i < \vartheta$, for some threshold $\vartheta$, and partitions the set $E$ into two subsets. These are the set $E_1$ of test-positive instances for which $x_i < \vartheta$ and the set $E_2$ of test-negative instances for which $x_i \geq \vartheta$, i.e., $\mathcal{P}^* =\{E_1, E_2\}$ (line \ref{algline:pctPartition}).

The procedure is recursively repeated on the subsets $E_i$ to construct the sub-trees (line~\ref{algline:recursion}) until a stopping criterion is satisfied (e.g., the minimal number of examples in a leaf is reached or the heuristic score no longer changes, etc.). In turn, a leaf node is created and its prototype is computed (line \ref{algline:leaf}). In the global MTR setting, the prototype is a vector of average target values of the examples in the leaf. The prototypes are used for prediction.

 \begin{algorithm}[b]
    \caption{Random Forest of PCTs ($\datatrain$, $\mathrm{M}$, $f$)}\label{alg:pct}
     \begin{algorithmic}[1]
        \STATE{$\forest = \emptyset$}
        \FOR{$m = 1, 2, \dots, \mathrm{M}$}
            \STATE{$E = \mathit{bootstrapSample}(\datatrain)$} \label{algline:bSamle}
            \STATE{$\tree = \mathit{inducePCT(E,f)}$}\label{algline:PCTcall}
            \STATE{add $\tree$ to $\forest$}
        \ENDFOR
    \RETURN{$\forest$}
    \vskip0.5cm
    \ALOOP{$\mathit{inducePCT}(E, f)$}
        \STATE{$\features =$ random sample of $f$ features}\label{algline:pctS}
        \STATE $(t^*, \mathcal{P}^*) = \mathit{findBestTest}(E, \features)$\label{algline:pctPartition}
         \IF{$t^* \neq \mathit{none}$}
             \FOR{\textbf{each} $E_i \in \mathcal{P^*}$}
                \STATE $\mathit{subtree}_i = \mathit{inducePCT}(E_i, f)$\label{algline:recursion}
             \ENDFOR
             \STATE{\textbf{return} $\node(t^*,\;\bigcup_i \{\mathit{subtree}_i\})$\hfill // internal node}
         \ELSE
             \STATE{\textbf{return} $\leaf(\mathrm{Prototype}(E))$ \hfill // leaf node}\label{algline:pctEND}\label{algline:leaf}
         \ENDIF   \ENDALOOP
    \end{algorithmic}
    \end{algorithm}

Finally, the RF of PCTs algorithm outputs a set of PCTs, which predictions are combined (averaged per value) to obtain the final ensemble prediction. The reasons for using RF of PCTs are twofold: (i) its state-of the-art predictive performance \cite{Kocev13:jrnl}, and (ii) ability to calculate feature importance scores, i.e., ranking of the features w.r.t. their importance for the target variables.

Namely, random forests can measure how much each feature contributed to the quality of the predictive model. For this purpose, we used the Genie3 algorithm \cite{genie3}, for which the motivation is the following. If a relevant feature $x_i$ is part of a test $x_i < \vartheta$, then the heuristic score $h^*$ (reduction of the variance) of this split is high. Additionally, the features that appear in the tests of nodes at lower depths, e.g., in the root, influence more examples compared to the ones appearing deeper in the tree, so the former are intuitively more important. Therefore, the Genie3 importance score is defined as
\begin{equation}\label{eqn:genie3}
\importance_\text{GENIE3}(x_i) = \frac{1}{|\forest|} \sum_{\tree \in \forest} \sum_{\node \in \tree(x_i)} h^*(\node)|E(\node)|\text{,}
\end{equation}
where $\tree(x_i)$ is the set of nodes in the tree $\tree$ in which $x_i$ is part of the test,
$h^*(\node)$ is the value of the variance reduction function in the node $\node{}$ and $|E(\node)|$ is the number of examples that come to the node $\node$.

Further in the paper, we denote the local and the global version of Random Forest PCTs for multi-target regression with \rfstr{} and \rfmtr{}, respectively. 

\subsection{Gradient Boosted Trees}

Gradient boosting~\cite{Friedman2001} refers to a class of boosting ensemble methods \cite{hastie2013elements} which aims at learning a set of predictive models focusing on difficult observations in the data. In contrast to Random Forest ensembles, that first learn the ensemble constituents from random parts of the data set and in turn combine their outputs, boosting ensembles are constructed iteratively. At each boosting iteration a new weak base model that corrects the error made by the ensemble thus far is learned and added to the set creating a stronger model at each step. Typically, such weak models have simple structure and thus perform slightly better than random models.

In general, boosting methods differ in the type of base models they employ and how the learning is performed. The former is task related, where typical base models considered include: logistic regressors (for classification tasks), linear regressors (for regression tasks) or decision trees (for both classification and regression tasks). Regarding how the learning is performed, boosting base models are either constructed to focus on hard examples identified in previous iterations \cite{FreundSchapire1997} or by minimizing the empirical risk via steepest gradient descent \cite{Friedman2001}. In this paper we focus on the latter category, referred to as Gradient Boosting.

An outline of the gradient boosting algorithm for constructing ensembles with decision trees is presented in Alg.~\ref{alg:GBT}. The algorithm takes three inputs: (1) a training data set $\datatrain$, (2) number of boosting iterations $\mathrm{M}$, (3) a loss function $\mathcal{L}$ and (4) a learning rate $\eta$. First, an initial default (weak) model is learned on the whole data set that minimizes the loss function $\mathcal{L}$. Given that the gradient boosted method aims at optimizing the loss function, it is important that $\mathcal{L}$ should be convex and differentiable. A typical choice of $\mathcal{L}$ is $L_{2}$ square-loss function \cite{hastie2013elements}. In turn, gradient boosting at each step $m$ learns a new model on the pseudo-residuals, i.e. the discrepancy value between the true and the predicted value of the ensemble in the previous iterations (line~\ref{algline:gbREZ}).

However, such straightforward approach can very easily overfit to the training data. In order to address the problem of overfitting, more sophisticated gradient boosting methods implement two different mechanisms: a learning rate and random data sampling procedure. The former regulates the influence of the prediction of each subsequent model added in the ensemble set (line~\ref{algline:gbLR}). The latter, referred to as Stochastic Gradient Boosting \cite{Friedman2002}, employs additional random data sampling procedure: Each model is learned and evaluated on different random subsamples of the training data (line~\ref{algline:gbSample}). 

\begin{algorithm}[ht]
    \caption{Stohastic Gradient Boosted Trees ($\datatrain$, $\mathrm{M}$, $\mathcal{L}$, $\eta$)}\label{alg:GBT}
     \begin{algorithmic}[1]
        \STATE{$GBT_{\emptyset} = \mathit{defaultModel}(\datatrain,\mathcal{L})$}
        \FOR{$m = 1, 2, \dots, \mathrm{M}$}
            \STATE{$E = \mathit{randomSample}(\datatrain)$} \label{algline:gbSample}
            \vskip0.3cm
            \STATE{$\mathscr{R_\text{m}}=\displaystyle\frac{\partial\mathcal{L}(E,GBT_{m-1}(E))}{\partial GBT_{m-1}}$}\hfill // compute pseudo-residuals  \label{algline:gbREZ}
             \vskip0.3cm
            \STATE{$\tree = \mathit{induce}(\mathscr{R_\text{m}},\mathcal{L})$}
            \STATE{$GBT_{m}=GBT_{m-1}+\eta\tree$}\label{algline:gbLR}
        \ENDFOR
    \RETURN{$GBT$}
     \end{algorithmic}
    \end{algorithm}

In this study, we employ XGBoost \cite{Chen2016} -- a recent efficient implementation of Stochastic Gradient Boosting \cite{Friedman2002} that employs regression trees as proposed in \cite{Breiman84:book} as base models. In the paper, we denote the XGBoost ensembles with \xgb{}. 

\section{Experimental Setup}\label{sec:exp-setup}

\subsection{Parameter Instantiation}\label{sec:paramethers}
\emph{Granularity.} The data granularity is defined by the length $\deltaT$ of the time interval that corresponds to one example in the data set. We construct the predictive models using data sets with $\deltaT \in \{1, 5, 10, 15, 30, 60\}$ (measured in minutes) where $\deltaT = 1$ corresponds to the data set used in \cite{smc:2017}.

\emph{Historical features}. For the data set with finest granularity $\deltaT = 1$, we consider the following numbers $H$ of historical intervals from \eqref{eqn:historic-feats}: $H\in \{4, 16, 32, 64, 128\}$. Consequently, the historical time span ranges from $4$ minutes to $128$ minutes. The choices of the historical intervals in the data sets with courser granularity are presented in Table~\ref{tab:histH}. In total, these values yield $35$ features in the data set with $\deltaT \leq 30$ and $21$ features data set with  $\deltaT = 60$. 

\begin{table}[!htb]
\centering
\caption{Values of the number of historic intervals $H$ and the corresponding historic time spans, for different granularities. $\deltaT$}\label{tab:histH}
\begin{tabular}{c| c c}
\hline
$\deltaT$ & values of $H$ & time spans \\ \hline
$1$ & $\{4, 16, 32, 64, 128\}$ & $\{4, 16, 32, 64, 128\}$ \\
$5$ & $\{1, 3, 6, 13, 25 \}$   & $\{5, 15, 30, 65, 125\}$ \\
$10$ & $\{1, 2, 3, 6, 13\}$ & $\{10, 20, 30, 60, 130\}$ \\
$15$ & $\{1, 2, 3, 4, 9\}$ & $\{15, 30, 45, 60, 135\}$ \\
$30$ & $\{1, 2, 3, 4, 5\}$ & $\{30, 60, 90, 120, 150\}$ \\
$60$ & $\{1, 2, 3\}$ & $\{60, 120, 180\}$ \\
\hline
\end{tabular}
\end{table}

\emph{Random Forest parameters.} To constrain the size of the trees in the Random Forests, we specify a minimal number of examples in the leaves $\instancesLeaf$ for each tree. Since the number of instances in the data sets is inversely proportional to $\deltaT$, we set the minimal number of instances for the $\deltaT = 1$ experiments to $500$, while for the others we set them to $\instancesLeaf = 500 / \deltaT$. Additionally, we set the total number of trees in the random forests to $200$, where one quarter of the features is considered at every split when growing the trees, i.e., $f = 0.25|\features|$ in Alg.~\ref{alg:pct}.

\emph{XGBoost parameters.} Analogously, to constrain the size of the trees we set maximal depth of each tree in the ensemble to $11$. The learning rate parameter is set to $0.1$. Additionally, to address potential over-fitting issues, for every boosting iteration $60\%$ of the features and $60\%$ of the examples are randomly chosen for training. The maximum number of boosting iteration (ensemble constituents) is set to $200$ with an early-stop option, i.e., if the newly added trees in the ensemble do not improve the performance of the ensemble over five consecutive boosting iterations the algorithm stops.

\subsection{Evaluation procedure}
The data set $\dataset$ consists of examples $(\bm{x}, \bm{y})$ where $\bm{x}$ is a vector of feature values (features are described in Sec.~\ref{sec:feats}),
and $\bm{y}$ is a vector of $33$ target values, i.e., the electrical currents trough the heaters and coolers.

The data set is divided into two parts: $\datatrain$ that describes the state of the spacecraft throughout the first three Martian years, and $\datatest$ that describes the state of the spacecraft throughout the fourth Martian year. All predictive models, i.e., the approximations $\hat{\bm{y}}: \bm{x}\mapsto \hat{\bm{y}}(\bm{x})$ of the true mappings $\bm{y}: \bm{x} \mapsto \bm{y}(\bm{x})$, were learned on $\datatrain$. The $i$-th component of vectors $\hat{\bm{y}}(\bm{x})$ and $\bm{y}(\bm{x})$, i.e., the predicted and true value for $i$-th target, are denoted by $\hat{y}_i(\bm{x})$ and $y_i(\bm{x})$.

The quality of the predictions $\hat{\bm{y}}(\bm{x})$ is evaluated on a separate test set $\datatest$, not used for learning the models. It is measured in terms of the average root mean squared error $\armse{}$, defined via the root mean squared errors of each target variable, as follows:
\begin{eqnarray}
\armse{}(\hat{\bm{y}}) &=& \sqrt{\frac{1}{T}\sum_{i = 1}^T \rmse{}^2(\hat{y}_i)}\label{eqn:armse}\\
\rmse{}(\hat{y}_i) &=& \sqrt{\frac{1}{|\datatest|}\sum_{(\bm{x}, \bm{y})\in \datatest} \left(y_i(\bm{x}) - \hat{y}_i(\bm{x})\right)^2}\label{eqn:rmse}
\end{eqnarray}

where $|\datatest|$ denotes the size of $\datatest$ and $T = 33$ is the number of target variables. 

We also compare the machine learning methods in terms of their time efficiency (for constructing a predictive model). By doing so, we estimate the trade-off between the predictive performance of the models and time needed for constructing them, in turn determining the optimal combination of time resolution in the data and machine learning method. The time efficiency refers to single-threaded runs of the algorithms, computed as follows. First, only an $\alpha$ portion of the ensemble is constructed where we measure the learning time $t_\alpha$. Subsequently, the total learning time is estimated as $t = t_\alpha / \alpha$. Such estimations were necessary, since some methods do not allow for single threaded runs (as reported in the next section).

\section{Results}
The goal of the experiments is to determine whether we can improve the efficiency of our approach for predicting thermal power consumption, while retaining good predictive performance. In particular, we evaluate tree different algorithms in terms of time efficiency (for learning a model) and predictive performance. 

First, we report on the running times of the three algorithms given training data with different granularity as well as the impact on their predictive performance. Next, we discuss different alternatives for improving both the efficiency and the predictive performance of the algorithms. Finally, we discuss the quality of the learned predictive models using feature importance diagrams. 

\subsection{Performance}\label{sec:performance}

\begin{figure}[!t]
\centering  
\includegraphics[width=0.6\textwidth]{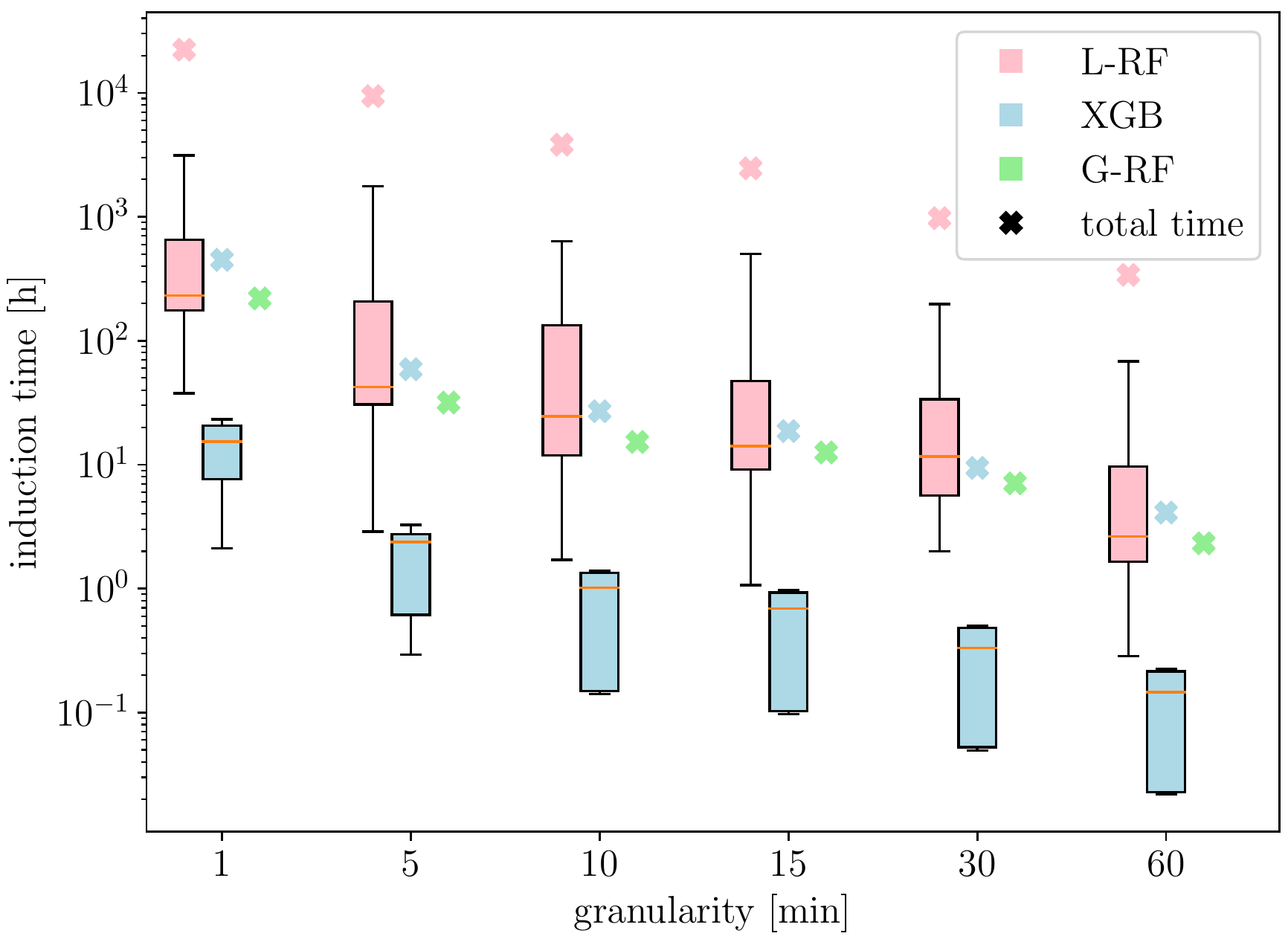}
\caption{Learning times for different algorithms and time granularity $\deltaT$. The crosses present the total learning time of the algorithm. Since \rfstr{} and \xgb{} build a model for each target (power line) separately, the distribution of learning times per power lines is additionally presented with the box plot. The per-target times add up to the total running time.}
\label{fig:granularity:time}
\end{figure}

We first focus on the learning times for different algorithms given different data granularity $\deltaT$. Figure~\ref{fig:granularity:time} presents the time needed for each algorithm to construct a model for each target as well as the total time to complete the task. 

As expected, in general, for all approaches the learning times decrease with the granularity $\deltaT$ increasing. Nevertheless, the \rfstr{} approach, that constructs an ensemble model for each of the $33$ target variables, has the longest learning time of approximately $22000$ hours ($\sim 2.5$ years). In particular, learning an \rfstr{} ensemble from the $\deltaT=1$ data set takes approximately $60$-times longer than constructing an \xgb{} ensemble which takes $450$ hours. 

While the constituents of an \xgb{} ensemble are considerably smaller than the ones of an $RF$, constructing a \xgb{} ensemble still takes approximately twice as much time as constructing a \rfmtr{} ensemble ($220$ hours). Note that, the constituents of both types of $RF$ ensembles can be constructed in parallel, thus additionally reducing the learning time for \rfstr{} and \rfmtr{} by a factor of $100$. In contrast, the most resource friendly is the $\deltaT = 60$ data set,  where the estimated learning times of \rfstr{}, \xgb{} and \rfmtr{} are $330$, $4.1$ and $2.3$ hours respectively. 

\begin{figure}[!b]
\centering
\includegraphics[width=0.6\textwidth]{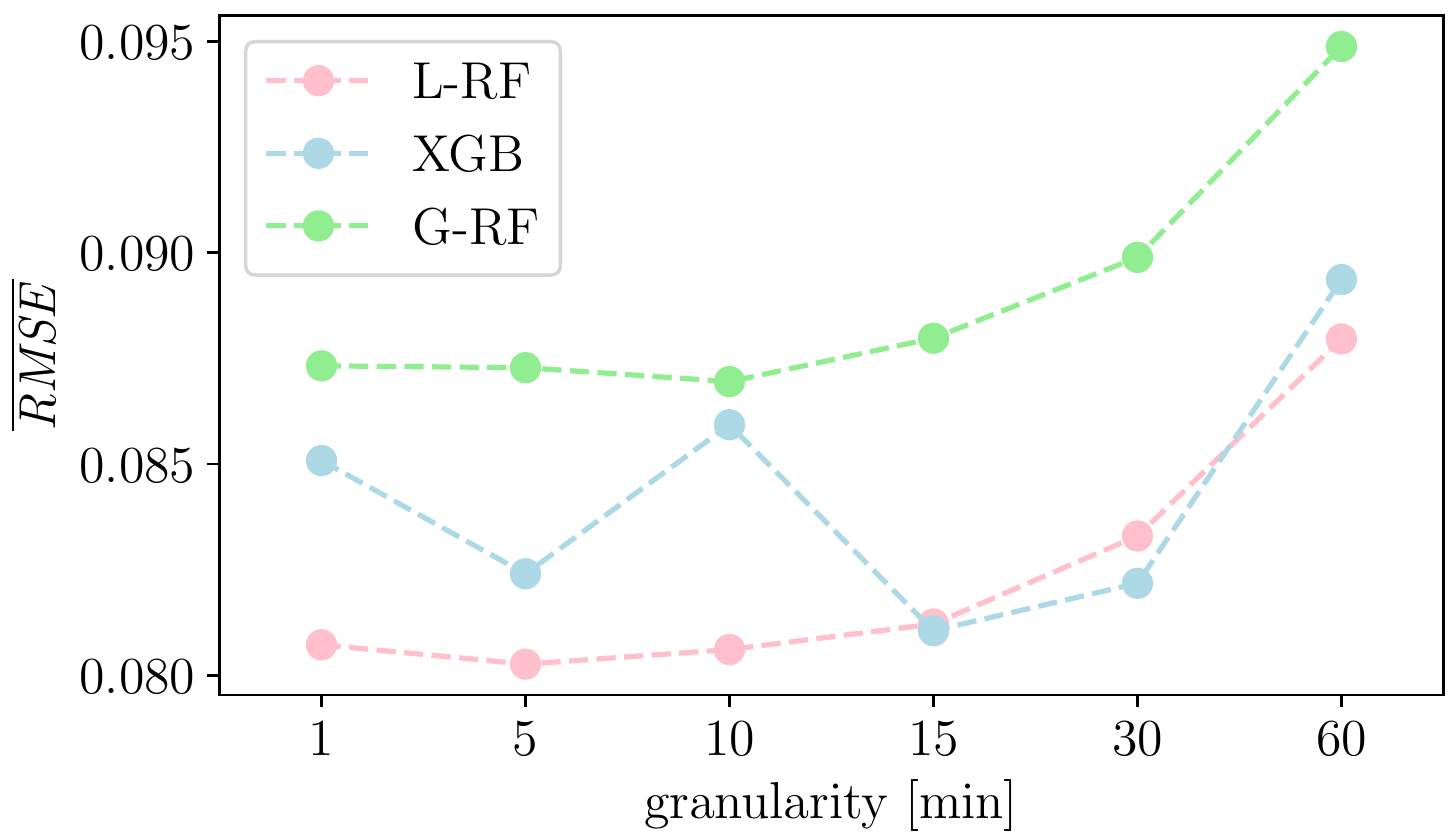}
\caption{Predictive performance of the three ensemble methods given data with different granularity.}\label{fig:granularity:rmse}
\end{figure}

Next, Figure~\ref{fig:granularity:rmse} presents the impact of the data granularity on the predictive performance of the models constructed with the three different methods. Note that there is a trade-off between learning-time efficiency and predictive performance. In particular, the ability of \rfmtr{} to construct predictive models in the shortest amount of time comes at a cost of decreased predictive performance compared to the other two methods. In this context, \xgb{} performs best in two cases ($\deltaT\in\{15, 30\}$) while being substantially more efficient than \rfstr{} which performs best in the remaining four cases ($\deltaT \in\{1, 5, 10, 60\}$).

The predictive error of both $RF$ methods, increases substantially with $\deltaT>15$. In the case of \xgb{}, however, the best predictive performance is obtained with the medium grained data sets $\deltaT\in \{15, 30\}$. Note that, the \rfstr{} trained on $\deltaT = 1$ data set (as used in \cite{smc:2017}) still has some competitive advantage in terms of predictive performance. However, similar (or slightly better) performance can be obtained either by learning from the $\deltaT\in \{5, 10\}$ which is considerably more efficient, or by constructing \xgb{} ensemble using the $\deltaT = 15$ data set.

\subsection{Further optimization}

So far, we reported on ensembles consisting of $200$ constituents. In general, the size of the ensemble has different effects on the quality of the predictions in the case of Random Forest ensembles and Boosting ensembles. In the former case, a general rule-of-thumb is that the predictive performance increases with the ensemble size, until it (effectively) saturates at some point. The reason for this is that every tree in RF is grown independently of the others. Thus, a RF ensemble may only be \emph{unnecessarily} large.

In contrast,
the boosting ensemble creation is different. Here, every additional tree focuses on minimizing the errors of the previously grown trees, therefore such ensembles have a tendency to overfit to the training data. As a consequence, the ensemble size can have substantial effect on the predictive performance.

\begin{figure}[b]
\centering
\includegraphics[width=0.7\textwidth]{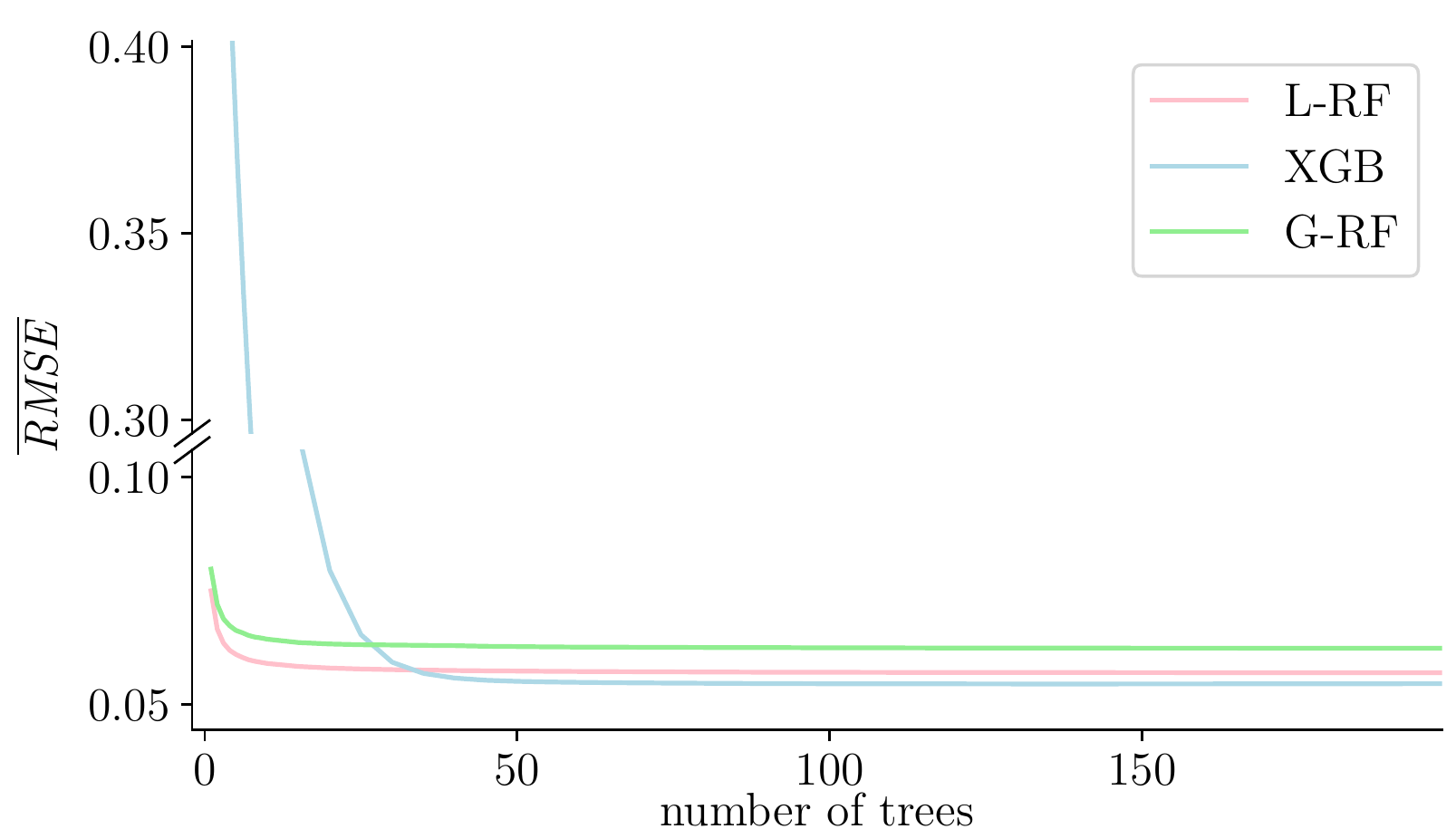}
\caption{The effect of the ensemble size on the performance of a Gradient Boosting ensemble (\xgb{}), a local Random Forest ensemble (\rfstr{}), and a global Random Forest ensemble (\rfmtr{}), evaluated with 5-fold cross validation on the $\deltaT=60$ data set.}
\label{fig:learningC}
\end{figure}

We conjecture that such artefacts are also present in the models evaluated thus far, and therefore we aim to further optimize the learning methods. 
To estimate the optimal number of trees in the ensembles, we take the $\deltaT = 60$ data set and perform 5-fold cross validation on $\datatrain$. More specifically, we randomly divide $\datatrain$ into five equally sized parts $P_i$, $1\leq i\leq 5$. We train ensembles with different sizes ($1$, \dots, $200$)
on every group of four parts, and estimate their performance on the remaining part not used for training. The average error of the five attempts (so called \emph{cross-validated} error) is the final score of a given ensemble. We perform the same procedures for the RF (\rfmtr) and \xgb{} ensembles.  

The results in Figure~\ref{fig:learningC} show that both types of methods can achieve good performance with considerably smaller ensembles. In particular, the RF method is able to achieve good performance very early (which does not drastically improve over time) due to accurate and deeper trees. On the other hand, as expected, \xgb{} starts with very poor performance which considerably improves after approximately $50$ iterations. 

Given these evidence, we once again evaluate the predictive performance of the three methods on the test data, however instead of constructing ensembles with $200$ trees we construct them with only $50$.

\begin{table}[t]
  \centering
  \caption{Predictive performance ($\armse{}$) of the three ensemble methods with 50 constituents.}
    \begin{tabular}{l|rrrrrr}
   \diagbox{Method}{\\ $\deltaT$}      & 1       & 5       & 10      & 15      & 30      & 60 \\
    \hline
    \rfstr{} & 0.0825 & 0.0804 & 0.0812 & 0.0808 & 0.0844 & 0.0876 \\
    \rfmtr{}   & 0.0865  &	0.0869  & 0.0881  & 0.0989  & 0.0899  & 0.0954 \\
    \xgb{}     & 0.0835  & 0.0796  & 0.0835  & \textbf{0.0785}  & 0.0816  & 0.0889 \\
    \end{tabular}%
  \label{tab:error50}%
\end{table}%

Table \ref{tab:error50} confirms our conjecture: The ensemble size in the case of RF methods (\rfstr{} and \rfmtr{}) has in general insignificant effect on the predictive performance. Note that, the Random Forest method is comprised of random trees, hence in some cases, like the \rfmtr{} constructed on the $\deltaT=15$ data set, small ensemble sizes hurt the predictive performance. Further analysis shows that in this case the performance stabilizes with at least $100$ constituents to $\armse=0.087$. In in the case of \xgb{}, the effect of the ensemble size is more prominent and constant. More specifically, the predictive performance of an \xgb{} ensemble learned on a data set with granularity $\deltaT=15$ yields the best overall performance we obtained so far. Note that obtaining such performance is also considerably faster than the one reported in our previous study, i.e., \rfstr{} learned on $\deltaT=1$ data set. 

\begin{figure}[b]
\centering
 \includegraphics[width=0.6\textwidth]{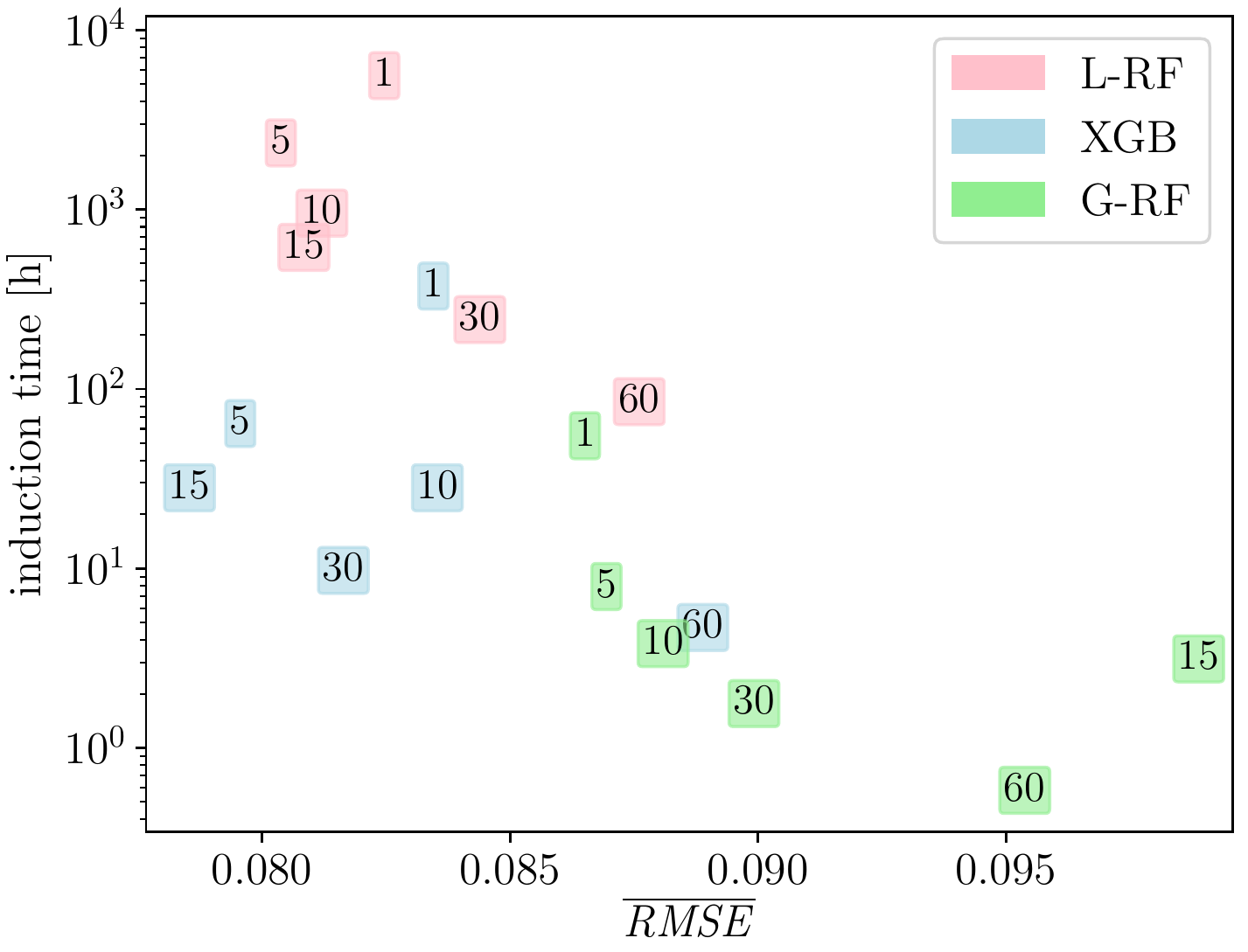}
\caption{The trade-off between the learning time and the predictive performance ($\armse{}$) of the three ensembles with $50$ : Colors of the rectangles denote the induction algorithm, while the numbers in the rectangles determine the time granularity $\deltaT$.}
\label{fig:times-rmse}
\end{figure}

Nevertheless, in terms of time efficiency, all ensemble methods benefit from the reduction of the ensemble size. In particular, on average the learning time is reduced by factor of $4$ in the case of RF ensembles and by a factor of almost $5$ in the case of boosting. Figure~\ref{fig:times-rmse} presents the results of the overall learning time and the predictive performance of all three methods with reduced ensemble size. 

More specifically, here we aim at identifying the optimal choice of algorithm given both criteria of predictive performance and time efficiency. Therefore, the optimal solution should be as close as possible to the lower left corner of the graph. A point $A$ in the graph is \emph{dominated} by another point $B$ in the graph if $B$ is better than $A$ in both criteria. For example, the performance of \rfstr{} on $\deltaT=10$ is dominated by the performance of \xgb{} with both  $\deltaT\in\{5,15\}$. The non-dominated points form a so called \emph{Pareto front}. In our case, the Pareto front consists of the two \xgb{} points  ($\deltaT\in\{15, 30\}$) and four \rfmtr{} points ($\deltaT \in\{5, 10, 30, 60\}$). All these points are considered optimal, unless we further specify our preferences over the criteria: If one aims at obtaining the fastest (but less accurate) solution one should consider \rfmtr{} ($\deltaT=60$). On the other hand, \xgb{} ($\deltaT=15$) yields the most accurate (but less efficient) solution.

\subsection{Ensemble of Ensembles}

\begin{figure}[t]
\centering
 \includegraphics[width=0.6\textwidth]{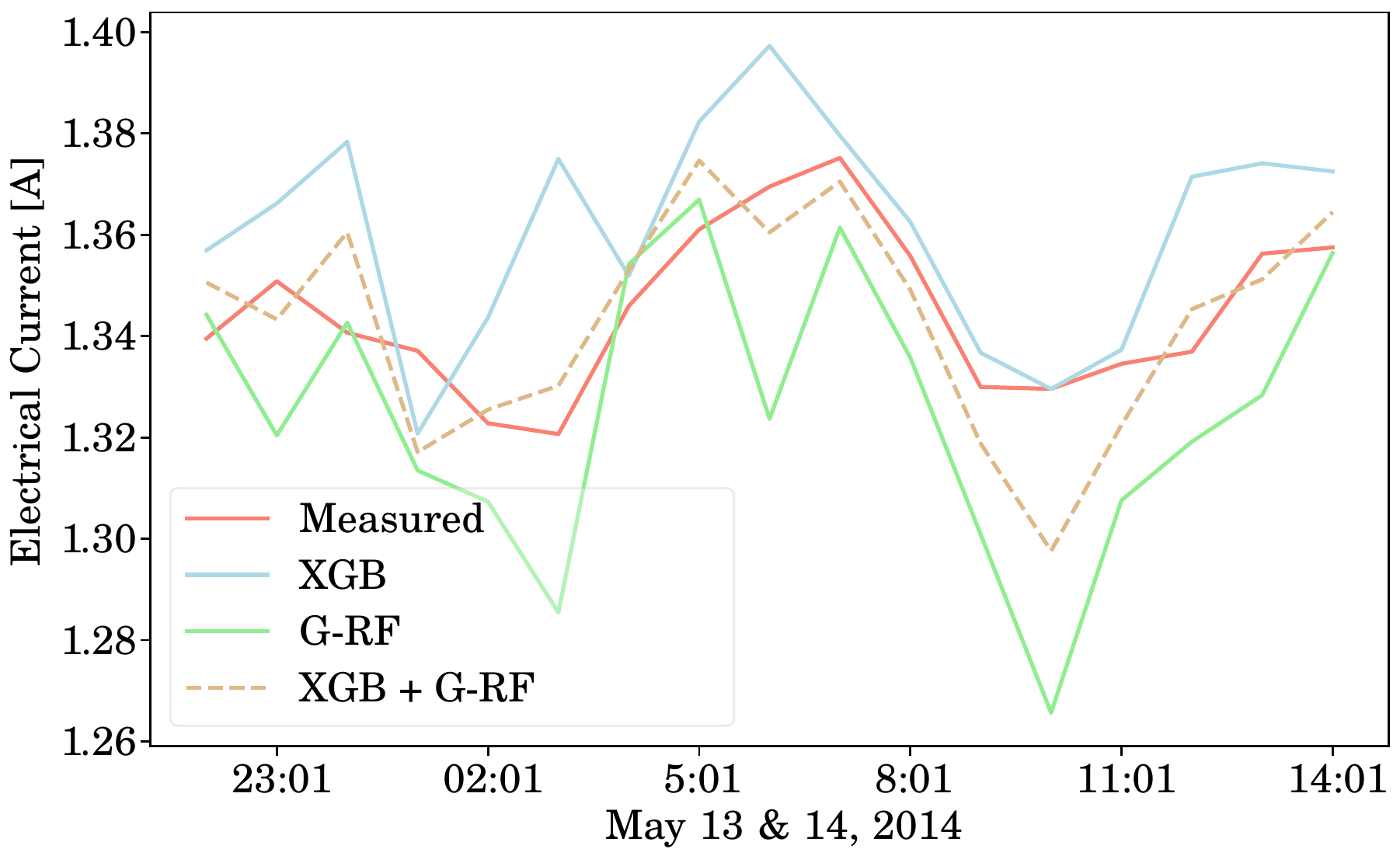}
\caption{Measured (red line) and predicted behavior on a sample time-period of 2 days for the 12th power line, obtained from \rfmtr{} and \xgb{} ensembles as well as their combination. }
\label{fig:enss}
\end{figure}

The performance of an ensembles is a consequence of the performance and diversity of its constituents. Moreover, ensembles usually perform better when compared to each individual constituent \cite{Hansen90:jrnl}. Given that in this study we consider different types of ensembles ({\it RFs} and \xgb{}) with good predictive performance, to further improve the overall performance we can also combine their outputs in an Ensemble of Ensembles (EoE). As a proof-of-principle, Figure~\ref{fig:enss} presents the predictions of \xgb{} and \rfmtr{} during sample time-period of 2 days for the 12th power line. We can see that, while the \xgb{} overestimates and \rfmtr{} underestimates the measured data (red line), their combination performs better.

Given the results from Figure~\ref{fig:times-rmse} we construct two EoEs. The member ensembles are trained on $\deltaT=15$ data set, since all results so far point to a good trade-off between efficiency and performance in this case. Moreover, both EoEs have one \xgb{} member combined either with \rfmtr{} (both methods are efficient) or \rfstr{} (both methods are accurate). 

The results are summed up in Table~\ref{tab:enss}. While the \xgb{}\&\rfmtr{} ensemble is very efficient to construct, its performance is only better than one of the members, \rfmtr{}. On the other hand, the \xgb{}\&\rfstr{} achieves the best performance overall of $\armse{} = 0.07779$. However, in practice, obtaining such an ensemble takes 10 times more than learning only a  \xgb{} ensemble that has practically similar performance ($\armse{} = 0.07853$).

\begin{table}[!htb]
\centering
\caption{Comparison of the two ensembles of ensembles (EoE) in terms of predictive performance ($\armse{}$) to the performance of the individual ensembles. All models are learned on the $\deltaT=15$ data set. }
\begin{tabular}{l| r}
 Ensembles &  $\armse{}$  \\
 \hline
 \xgb{}\&\rfmtr{} &  0.0803 \\
 \xgb{}\&\rfstr{} & \bf{0.0778} \\
 \hline
 \rfmtr{}(100) & 0.087 \\
 \rfstr{}(50) & 0.0808 \\
 \xgb{}(50) & 0.0785
\end{tabular}
\label{tab:enss}
\end{table}

\subsection{Feature Importance}

\begin{figure}[!b]
\centering
\subfloat[\rfmtr{}\label{fig:feats:mtr}]{
    \includegraphics[width=0.32\textwidth]{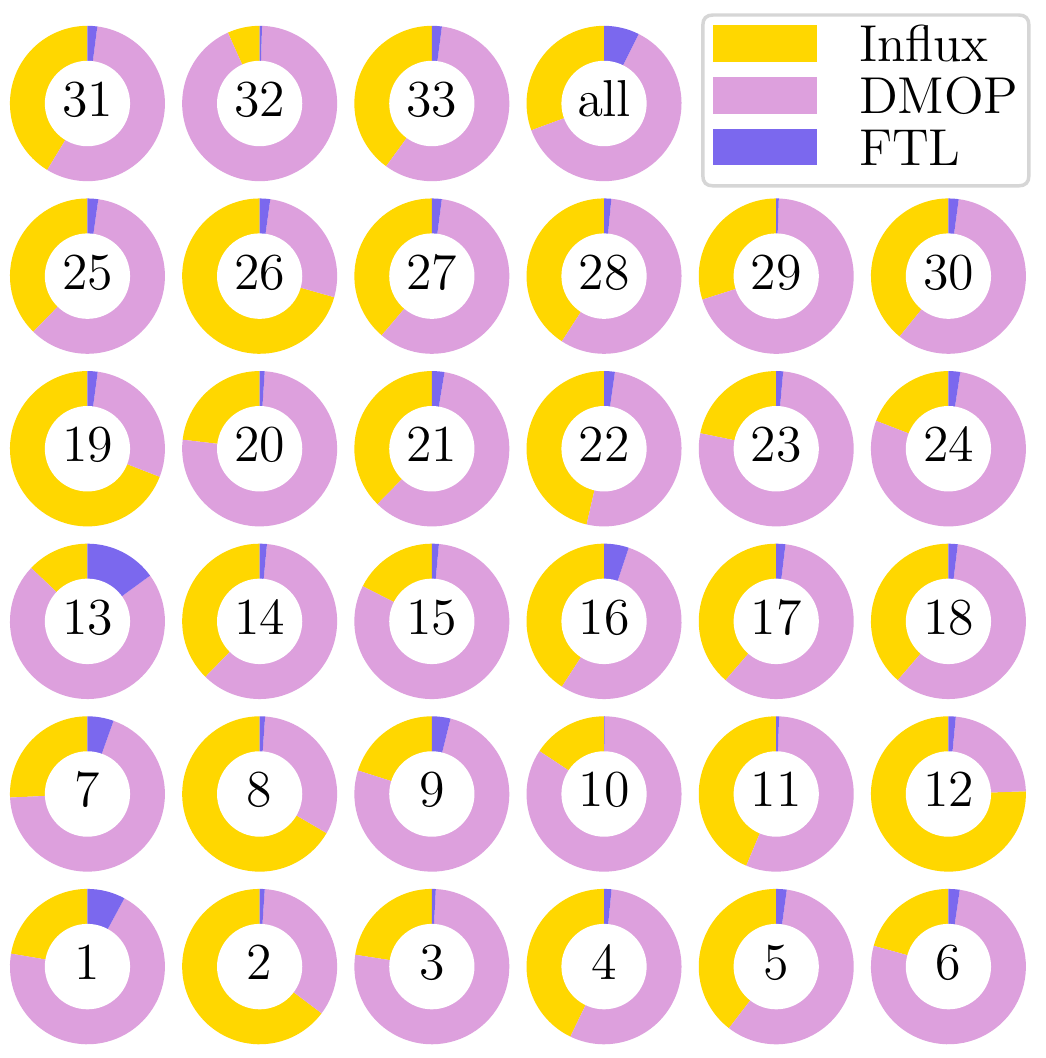}}
\hfill
\subfloat[\rfstr{}\label{fig:feats:str}]{
    \includegraphics[width=0.32\textwidth]{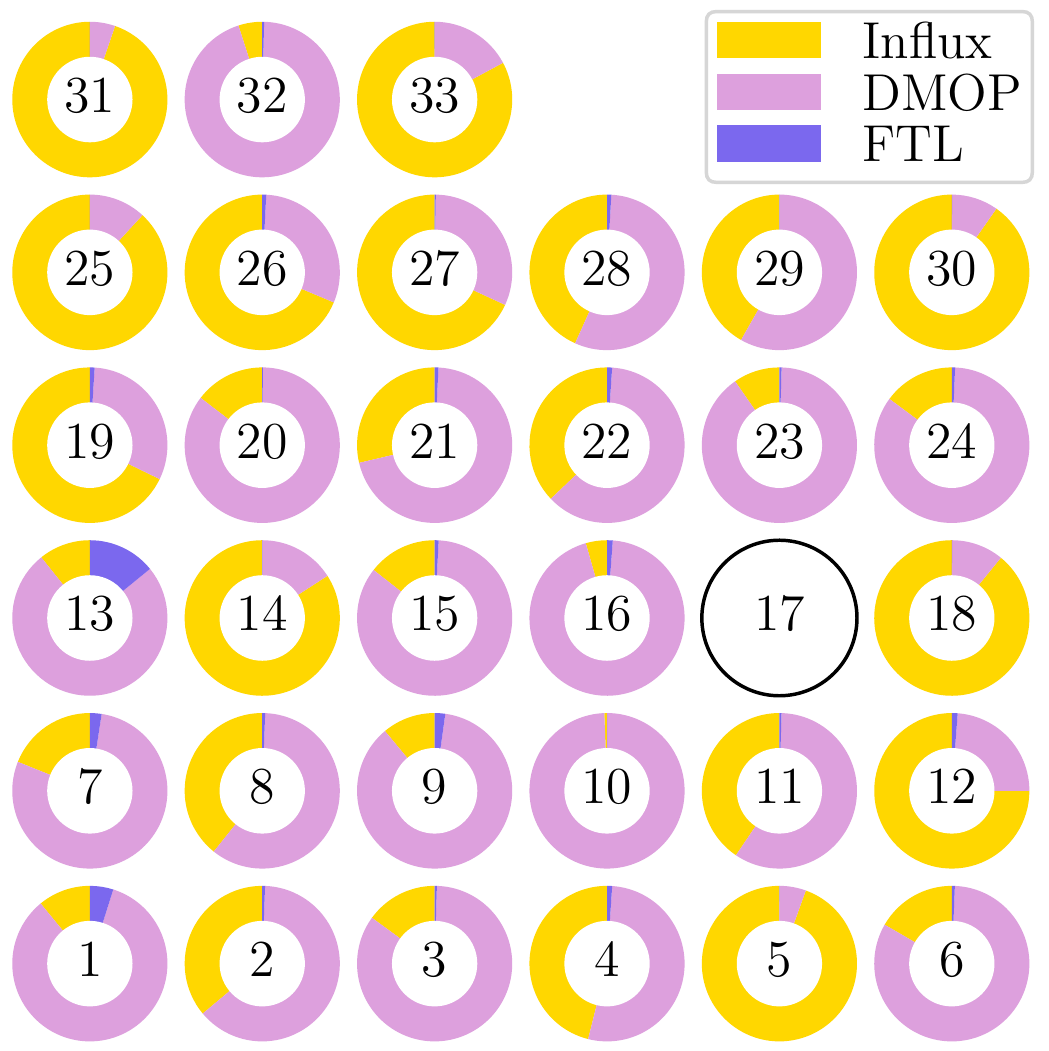}}
\hfill
\subfloat[\xgb{}\label{fig:feats:xgboost}]{
    \includegraphics[width=0.32\textwidth]{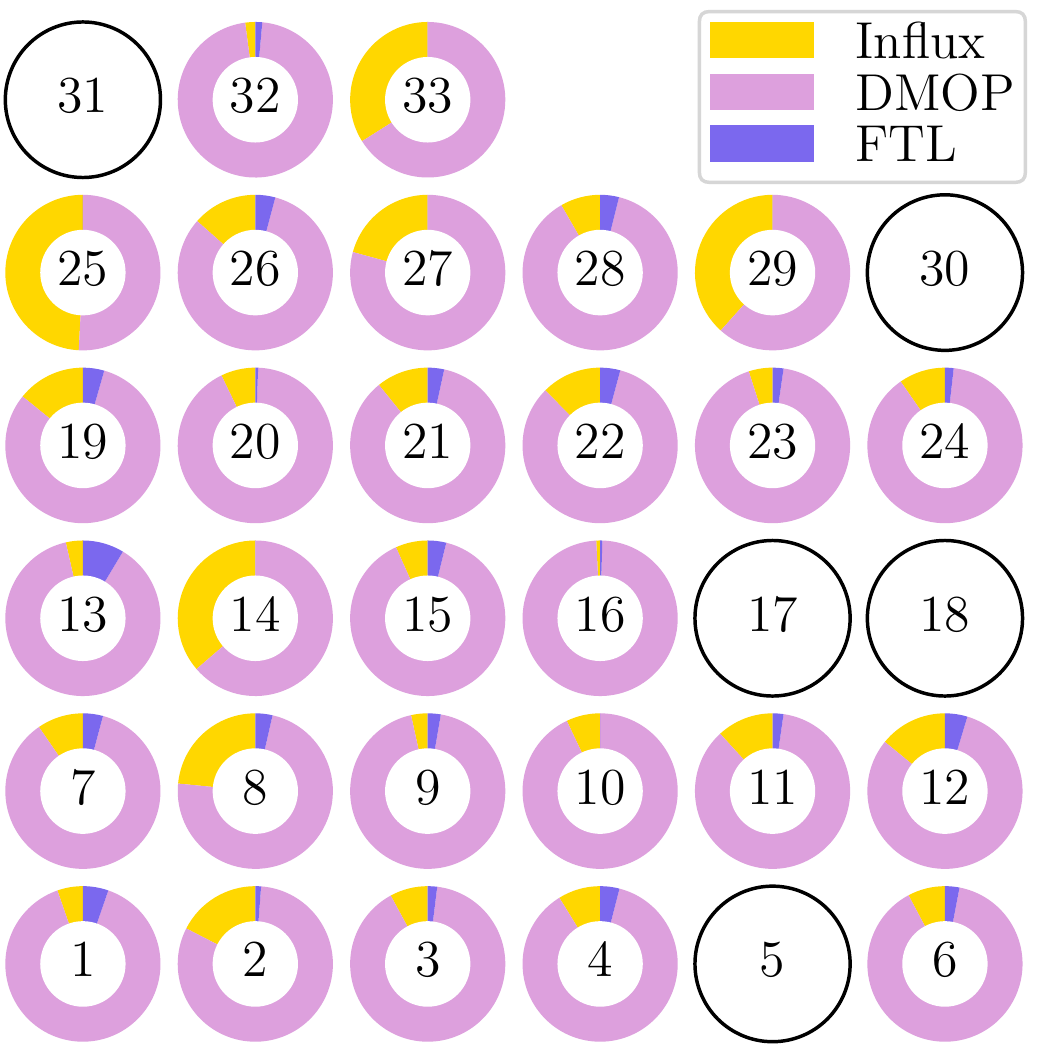}}
\caption{Distribution of different feature groups in the feature rankings produced by (a) \rfmtr, (b) \rfstr{} and (c) \xgb{} ensembles, for the $33$ power lines. In each of the individual diagrams, the presence of a given feature group type is proportional to the sum of Genie3 relevance over the features from the group. The empty diagrams denote ensembles constructed from constant models. The \rfmtr method allows for the additional \emph{global} feature importance denoted as \emph{all}.}
\label{fig:feats}
\end{figure}

In our last set of experiments, we assess the importance of the features obtained from the three methods. Typically, different features have different influence on the target variables, which in turn affects the predictive performance of the constructed models. Figure~\ref{fig:feats}, illustrates how different groups of features (energy influx, DMOP and FTL) influence the $33$ power consumers. The feature importance diagrams were calculated using the  $\deltaT=15$ data set. More specifically, in the case of \rfstr{} and \xgb{} (Figs.~\ref{fig:feats:str} and \ref{fig:feats:xgboost}), the proportions in the diagrams for each target were computed according to Eq.~\eqref{eqn:genie3}. On the other hand, in the case of \rfmtr{} (Figure~\ref{fig:feats:mtr}) where the model predicts all targets simultaneously, Eq.~\ref{eqn:genie3} leads only to the global feature importance (\emph{all} diagram). The per-target importance diagrams were computed from the local models, considering one target when computing the heuristic function. Note that, for some targets, the feature importance diagrams are blank since all ensemble constitutes in these cases are constant models (trees without internal nodes).

In the case of \rfmtr{}, the most important feature group overall is DMOP. In particular, in the majority of the power lines (27) their influence is at least $50\%$. The Energy influx features have a major influence on five power lines, while the FTL features have a considerable impact only on the 13th power line. Similarly, \rfstr{} finds the DMOP features as most important overall. However, as opposed to \rfmtr{}, here we can see find more power lines which are almost exclusively influenced by energy influx features. These differences in the computed feature importance between \rfmtr{} and \rfstr{} is a consequence of how the ensembles are constructed: While the former is able to capture the global phenomena across all power lines, the latter captures more detailed behavior that relates to each individual power line. 

On the other hand, \xgb{} mostly relates to the DMOP features. Compared to the other two methods, in this case their influence is considerably greater. In particular, the Energy Influx features have some significant importance ($>20\%$) on only 4 power lines, while the FTL features do not contribute greatly. Additionally, \xgb{} was not able to produce feature importance diagrams for 5 of the target variables.

\section{Conclusion}

In this paper, we propose a machine learning pipeline for predicting the power of the thermal subsystem on board the Mars Express spacecraft. More specifically, we propose novel solutions in the machine learning pipeline that focus on efficiently constructing predictive models of MEX's TPC, while still being able to maintain high predictive performance. More specifically, we employ state-of-the-art feature engineering approaches for transforming raw telemetry data, which in turn is used for constructing accurate predictive models with different machine learning methods. These solutions are the main contribution of our paper, since they considerably improve our competition-winning solution \cite{smc:2017} in two directions: efficiency and accuracy.

The proposed improvements in the pipeline consider (1) preparing training data with different time granularity, as well as (2) employing different machine learning methods for constructing accurate predictive models. Regarding the former, we carefully transformed the raw telemetry data at different time resolutions ($\deltaT\in\{1, 5, 10, 15, 30, 60\}$ min) which resulted in significant reductions of the size of the data sets used in the learning phase. Regarding the latter, we considered different state-of-the-art local and global machine learning ensemble methods for multi-target regression. These methods include: Local and Global Random Forests of Predictive Clustering Trees (\rfstr{} and \rfmtr{}) as well as Stochastic Gradient Boosted Trees (\xgb{}). We evaluated our proposed solutions on the task of predicting hourly values of the electric current through the $33$ thermal power consumers on board MEX for one Martian year, given raw telemetry data of three preceding Martian years.

In terms of time efficiency, our empirical study shows that the time resolution of the data has a significant impact on both the construction time of the predictive models as well as on their accuracy. The former is an expected result, given that coarser granularity yields a reduced data set and therefore shorter learning time. However, learning methods using coarser data usually yield less accurate models. The latter result though, provides a significant insight into this problem: Given a data set with moderate granularity ($\deltaT\in\{10,15\}$), all three methods are able to obtain models with comparable (or better) predictive performance, in substantially shorter time as compared to models learned on data set with finer granularity.

In terms of predictive performance, the local ensemble methods perform better than the global method. More specifically, in most cases, \rfstr{} and \xgb{} have comparable performance, with \xgb{} being slightly better. While both methods perform better than \rfmtr{}, the difference in performance is neither substantial nor significant. Note that learning a global model also takes considerably less time than learning a local model. In the same context, our results show that, while the size of the ensembles has a significant effect on the learning time, it can also improve the predictive performance. In particular, we showed that, with all methods we can obtain similar or better (\xgb{}) predictive performance with smaller ensembles. Moreover, we also demonstrated that by further combining the predictions of the different ensembles into an ensemble-of-ensembles, we additionally improve the predictive performance and obtain premium accuracy of $\armse_{\xgb{}\&\rfstr{}} = 0.07779$.

Finally, our feature importance analysis indicates that, for this particular problem of predicting thermal power consumption, the Detailed Mission Operations Plans (DMOP) have a significant role in the quality of the predictive models. Their importance is more prominent in the \xgb{} ensembles models than in the \textit{RF} ensembles, which also rely on the Energy Influx features when constructing a model. 

There are several directions to extend the work presented in this paper. Considering the data, note that DMOP information is available only after a certain command is executed on the spacecraft and its effect measured subsequently. This means that using these data for predicting longer time horizons is not possible. Moreover, given the findings of our paper, omitting them from the learning process might have a severe consequence on the performance of the predictive models. Therefore, an immediate continuation of the work presented here is to further optimize the constructed features as well as investigate different approaches for engineering new (informative) features. Finally, while the proposed methodology focuses on the thermal subsystem of the MEX spacecraft, it can also be readily applied to the other subsystems. Moreover, it can also be extended to other spacecraft such as the XMM Newton \cite{canio:spaceops}, Integral \cite{finn:spaceops} and ExoMars as well as rovers (such as Curiosity and ExoMars) exploring Mars. 

\section*{Acknowledgment}
We thank the Slovenian grid initiative \& the Academic and Research Network of Slovenia for support with the computer infrastructure. 
\section*{Funding}
We acknowledge the financial support of the Slovenian Research Agency (via the grants P2-0103, J4-7362, L2-7509 and the young researcher grants to MP and MB), as well as the European Commission via the grants HBP (The Human Brain Project) SGA1 and SGA2.
\section*{Competing interests}
The authors declare that no competing interests exist.

\ifCLASSOPTIONcaptionsoff
  \newpage
\fi

\bibliographystyle{IEEEtran}
%\bibliography{references}

\begin{thebibliography}{10}
	\providecommand{\url}[1]{#1}
	\csname url@samestyle\endcsname
	\providecommand{\newblock}{\relax}
	\providecommand{\bibinfo}[2]{#2}
	\providecommand{\BIBentrySTDinterwordspacing}{\spaceskip=0pt\relax}
	\providecommand{\BIBentryALTinterwordstretchfactor}{4}
	\providecommand{\BIBentryALTinterwordspacing}{\spaceskip=\fontdimen2\font plus
		\BIBentryALTinterwordstretchfactor\fontdimen3\font minus
		\fontdimen4\font\relax}
	\providecommand{\BIBforeignlanguage}[2]{{%
			\expandafter\ifx\csname l@#1\endcsname\relax
			\typeout{** WARNING: IEEEtran.bst: No hyphenation pattern has been}%
			\typeout{** loaded for the language `#1'. Using the pattern for}%
			\typeout{** the default language instead.}%
			\else
			\language=\csname l@#1\endcsname
			\fi
			#2}}
	\providecommand{\BIBdecl}{\relax}
	\BIBdecl
	
	\bibitem{Oroseieaar7268}
	R.~Orosei, S.~E. Lauro, E.~Pettinelli, A.~Cicchetti, M.~Coradini, B.~Cosciotti,
	F.~Di~Paolo, E.~Flamini, E.~Mattei, M.~Pajola, F.~Soldovieri, M.~Cartacci,
	F.~Cassenti, A.~Frigeri, S.~Giuppi, R.~Martufi, A.~Masdea, G.~Mitri,
	C.~Nenna, R.~Noschese, M.~Restano, and R.~Seu, ``Radar evidence of subglacial
	liquid water on mars,'' \emph{Science}, vol. 361, pp. 490--493, 2018.
	
	\bibitem{chicarro2004mars}
	A.~Chicarro, P.~Martin, and R.~Trautner, ``The {Mars} {Express} mission: An
	overview,'' in \emph{Mars Express: The Scientific Payload}, vol. 1240, 2004,
	pp. 3--13.
	
	\bibitem{lucas2017:mars}
	L.~Lucas and R.~Boumghar, ``Machine learning for spacecraft operations support
	- {The Mars Express Power Challenge},'' in \emph{Sixth International
		Conference on Space Mission Challenges for Information Technology, {SMC-IT
			2017}}, 2017, pp. 82--87.
	
	\bibitem{smc:2017}
	M.~Breskvar, D.~Kocev, J.~Levati\'{c}, A.~Osojnik, M.~Petkovi\'{c},
	N.~Simidjievski, B.~\v{Z}enko, R.~Boumghar, and L.~Lucas, ``Predicting
	thermal power consumption of the mars express satellite with machine
	learning,'' in \emph{2017 6th International Conference on Space Mission
		Challenges for Information Technology (SMC-IT)}, 2017, pp. 88--93.
	
	\bibitem{Kocev13:jrnl}
	D.~Kocev, C.~Vens, J.~Struyf, and S.~D\v{z}eroski, ``Tree ensembles for
	predicting structured outputs,'' \emph{Pattern Recognition}, vol.~46, no.~3,
	pp. 817--833, 2013.
	
	\bibitem{Chen2016}
	T.~Chen and C.~Guestrin, ``Xgboost: A scalable tree boosting system,'' in
	\emph{Proceedings of the 22Nd ACM SIGKDD International Conference on
		Knowledge Discovery and Data Mining}, ser. KDD '16, 2016, pp. 785--794.
	
	\bibitem{Friedman2002}
	J.~H. Friedman, ``Stochastic gradient boosting,'' \emph{Computational
		Statistics \& Data Analysis}, vol.~38, no.~4, pp. 367 -- 378, 2002.
	
	\bibitem{borchani}
	H.~Borchani, G.~Varando, C.~Bielza, and P.~Larra\~{n}aga, ``A survey on
	multi-output regression,'' \emph{Data Mining and Knowledge Discovery},
	vol.~5, no.~5, pp. 216--233, 2015.
	
	\bibitem{Spyromitros-Xioufis2016}
	E.~Spyromitros-Xioufis, G.~Tsoumakas, W.~Groves, and I.~Vlahavas,
	``Multi-target regression via input space expansion: treating targets as
	inputs,'' \emph{Machine Learning}, vol. 104, no.~1, pp. 55--98, 2016.
	
	\bibitem{hastie2013elements}
	T.~Hastie, R.~Tibshirani, and J.~Friedman, \emph{The Elements of Statistical
		Learning: Data Mining, Inference, and Prediction}, ser. Springer Series in
	Statistics.\hskip 1em plus 0.5em minus 0.4em\relax Springer New York, 2013.
	
	\bibitem{Witten2005}
	I.~H. Witten and E.~Frank, \emph{Data Mining: {P}ractical Machine Learning
		Tools and Techniques}.\hskip 1em plus 0.5em minus 0.4em\relax Morgan
	Kaufmann, 2005.
	
	\bibitem{Bakir2007}
	G.~H. Bak{\i}r, T.~Hofmann, B.~Sch{\"o}lkopf, A.~J. Smola, B.~Taskar, and
	S.~V.~N. Vishwanathan, \emph{Predicting structured data}, ser. Neural
	Information Processing.\hskip 1em plus 0.5em minus 0.4em\relax The MIT Press,
	2007.
	
	\bibitem{McGovern2011}
	A.~McGovern and K.~L. Wagstaff, ``Machine learning in space: extending our
	reach,'' \emph{Machine Learning}, vol.~84, no.~3, pp. 335--340, 2011.
	
	\bibitem{Zhenping:smc}
	Z.~Li, ``Machine learning in spacecraft ground systems,'' in \emph{2017 6th
		International Conference on Space Mission Challenges for Information
		Technology (SMC-IT)}, 2017, pp. 76--81.
	
	\bibitem{Yairi}
	T.~Yairi, Y.~Kawahara, R.~Fujimaki, Y.~Sato, and K.~Machida,
	``Telemetry-mining: A machine learning approach to anomaly detection and
	fault diagnosis for space systems,'' in \emph{Proceedings - SMC-IT 2006: 2nd
		IEEE International Conference on Space Mission Challenges for Information
		Technology}, vol. 2006, 07 2006, pp. 476--483.
	
	\bibitem{Munoz:smc}
	M.~Mu\~{n}oz, Y.~Yue, and R.~Weber, ``Telemetry anomaly detection system using
	machine learning to streamline mission operations,'' in \emph{2017 6th
		International Conference on Space Mission Challenges for Information
		Technology (SMC-IT)}, 2017, pp. 70--75.
	
	\bibitem{canio:spaceops}
	G.~De~Canio, T.~Godard, R.~Boumghar, and U.~Weissmann, ``Optimization of the
	battery usage during eclipses using a machine learning approach,'' in
	\emph{15th International Conference on Space Operations}, Marseille, France,
	2018.
	
	\bibitem{Hernandez:smc}
	A.~C. Hern\'{a}ndez, C.~G\'{o}mez, J.~Crespo, and R.~Barber, ``Adding
	uncertainty to an object detection system for mobile robots,'' in \emph{2017
		6th International Conference on Space Mission Challenges for Information
		Technology (SMC-IT)}, 2017, pp. 7--12.
	
	\bibitem{Giusti:forest-paths}
	A.~Giusti, J.~Guzzi, D.~C. Cire\c{s}an, F.-L. He, J.~P. Rodr\'{i}guez,
	F.~Fontana, M.~Faessler, C.~Forster, J.~Schmidhuber, D.~Di~Caro,
	Gianni~Scaramuzza, and L.~M. Gambardella, ``A machine learning approach to
	visual perception of forest trails for mobile robots,'' \emph{IEEE Robotics
		and Automation Letters}, vol.~1, no.~2, pp. 661--667, 2016.
	
	\bibitem{finn:spaceops}
	T.~J. Finn, R.~Boumghar, J.~Martinez, and A.~Georgiadou, ``Machine learning
	modeling methods for radiation belts profile predictions,'' in \emph{15th
		International Conference on Space Operations}, Marseille, France, 2018.
	
	\bibitem{boumghar:spaceops}
	R.~Boumghar, L.~Lucas, and A.~Donati, ``Machine learning in operations for the
	mars express orbiter,'' in \emph{15th International Conference on Space
		Operations}, Marseille, France, 2018.
	
	\bibitem{Breiman01:jrnl}
	L.~Breiman, ``Random forests,'' \emph{Machine Learning}, vol.~45, no.~1, pp.
	5--32, 2001.
	
	\bibitem{blockeel-top-down-proc-1998}
	H.~Blockeel, L.~De~Raedt, and J.~Ramon, ``Top-down induction of clustering
	trees,'' in \emph{The 15th International Conference on Machine
		learning}.\hskip 1em plus 0.5em minus 0.4em\relax Morgan Kaufmann, San
	Francisco, CA, 1998, pp. 55--63.
	
	\bibitem{Breiman84:book}
	L.~Breiman, J.~Friedman, R.~Olshen, and C.~J. Stone, \emph{Classification and
		Regression Trees}.\hskip 1em plus 0.5em minus 0.4em\relax Chapman \&
	Hall/CRC, 1984.
	
	\bibitem{Blockeel98:phd}
	H.~Blockeel, ``Top-down induction of first order logical decision trees,''
	Ph.D. dissertation, Katholieke Universiteit Leuven, Leuven, Belgium, 1998.
	
	\bibitem{genie3}
	V.~A. Huynh-Thu, A.~Irrthum, L.~Wehenkel, and P.~Geurts, ``Inferring regulatory
	networks from expression data using tree-based methods,'' \emph{PLOS ONE},
	vol.~5, no.~9, pp. 1--10, 09 2010.
	
	\bibitem{Friedman2001}
	J.~H. Friedman, ``Greedy function approximation: A gradient boosting machine,''
	\emph{The Annals of Statistics}, vol.~29, no.~5, pp. 1189--1232, 2001.
	
	\bibitem{FreundSchapire1997}
	Y.~Freund and R.~E. Schapire, ``A decision-theoretic generalization of on-line
	learning and an application to boosting,'' \emph{Journal of Computer and
		System Sciences}, vol.~55, no.~1, pp. 119--139, 1997.
	
	\bibitem{Hansen90:jrnl}
	L.~K. Hansen and P.~Salamon, ``Neural network ensembles,'' \emph{IEEE
		Transactions on Pattern Analysis and Machine Intelligence}, vol.~12, pp.
	993--1001, 1990.
	
\end{thebibliography}

\end{document}